\documentclass[acmlarge]{acmart}

\usepackage{color}
\usepackage{amsthm}
\newtheorem{definition}{Definition} 

\usepackage{makecell}
\usepackage{multicol}
\usepackage{multirow}
\AtBeginDocument{%
  }

\setcopyright{acmlicensed}
\copyrightyear{2024}
\acmYear{2024}
\acmDOI{XXXXXXX.XXXXXXX}

\acmJournal{POMACS}
\acmVolume{37}
\acmNumber{4}
\acmArticle{111}
\acmMonth{8}

\DeclareMathOperator*{\argmin}{arg\,min}

\newcommand{\red}[1]{\textbf{\textcolor{red}{#1}}}
\newcommand{\writer}[1]{\textbf{\textcolor{red}{Author-{#1}}:}}
\begin{document}

\title{Evaluating and Improving Robustness in Large Language Models: A Survey and Future Directions}

\author{Kun Zhang}
\email{zhang1028kun@gmail.com}
\orcid{0000-0002-0743-9003}
\authornotemark[1]
\affiliation{%
  \institution{Hefei University of Technology}
  \city{Hefei}
  \state{Anhui}
  \country{China}
}

\author{Le Wu}
\affiliation{%
  \institution{Hefei University of Technology}
  \city{Hefei}
  \state{Anhui}
  \country{China}}
\email{lewu.ustc@gmail.com}

\author{Kui Yu}
\affiliation{%
  \institution{Hefei University of Technology}
  \city{Hefei}
  \state{Anhui}
  \country{China}
}
\email{yukui@hfut.edu.cn}


\author{Guangyi Lv}
\affiliation{%
 \institution{AI Laboratory, Lenovo Research}
 \city{Beijing}
 \state{Beijing}
 \country{China}
}
\email{gylv@mail.ustc.edu.cn}

\author{Dacao Zhang}
\affiliation{%
 \institution{Hefei University of Technology}
  \city{Hefei}
  \state{Anhui}
  \country{China}
}
\email{zhdacao@gmail.com}





\renewcommand{\shortauthors}{Zhang et al.}

\newcommand{\robust}{\emph{LLM Robustness}}
\newcommand{\mlrobust}{ML Robustness}
\newcommand{\github}{\url{https://github.com/zhangkunzk/Awesome-LLM-Robustness-papers}}
\newcommand{\tool}{\emph{Easy Search}}
\newcommand{\eai}{Embodied AI}

\begin{abstract}
  Large Language Models~(LLMs) have gained enormous attention in recent years due to their capability of understanding and generating natural languages. 
  With the rapid development and wild-range applications (e.g., Agents, Embodied Intelligence), the robustness of LLMs has received increased attention. 
  As the core brain of many AI applications, the robustness of LLMs requires that models should not only generate consistent contents, but also ensure the correctness and stability of generated content when dealing with \textit{unexpeted application scenarios}~(e.g., toxic prompts, limited noise domain data, out-of-distribution~(OOD) applications, etc).
  In this survey paper, we conduct a thorough review of the robustness of LLMs, aiming to provide a comprehensive terminology of concepts and methods around this field and facilitate the community. 
  Specifically, we first give a formal definition of LLM robustness and present the collection protocol of this survey paper. 
  Then, based on the types of perturbated inputs, we organize this survey from the following perspectives: 
  1) \textit{Adversarial Robustness}: tackling the problem that prompts are manipulated intentionally, such as noise prompts, long context, data attack, etc;   
  2) \textit{OOD Robustness}: dealing with the unexpected real-world application scenarios, such as OOD detection, zero-shot transferring, hallucinations, etc;
  3) \textit{Evaluation of Robustness}: summarizing the new evaluation datasets, metrics, and tools for verifying the robustness of LLMs.   
  After reviewing the representative work from each perspective, we discuss and highlight future opportunities and research directions in this field.
  Meanwhile, we also organize related works and provide an easy-to-search project~(\github) to support the community.
  

\end{abstract}



\received{20 August 2024}

\maketitle

\section{Introduction}
\label{s:intro}
Large Language Models~(LLMs), such as GPT-4~\cite{achiam2023gpt}, Llama~\cite{touvron2023llama}, Mistral~\cite{jiang2024mixtral}, and DeepSeek~\cite{liu2024deepseek} have achieved impressive performance over enormous areas (e.g., conversation, code, education, etc). 
Empowered by billions of parameters and vast training corpora, LLMs are capable of demonstrating emergent abilities~\cite{wei2022emergent} in comprehension, memorizing, and reasoning with human instructions. 
Moreover, AI agents and Embodied AI~(EAI)~\cite{paolo2024call} have treated LLMs as the brain of central scheduling, which are expected to revolutionize existing workflows~\cite{bellas2023ai,imani2023mathprompter,lan2024teachers}.
All of these applications place high demands on the reliability of LLMs, such as generating consistency contents when facing noising inputs and maintaining performance when dealing with domain shifting, which can be coined as the \textit{robustness problem} of LLMs. 
As one of the core components in Artificial Intelligence~(AI), the robustness of an algorithm has been defined as the \textit{insensitivity of a model’s performance to miscalculations of its parameters}~\cite{nobandegani2019robustness,zhang2020interpreting}. 
Generally, robustness in Machine Learning~(ML) focuses on natural (non-adversarial) perturbations $P_{train}(X, Y) \neq P_{text}(X, Y)$(i.e., different data distributions) and adversarial perturbations $(X,Y)\rightarrow (X+\delta, Y)$ (i.e., perturbations $\delta$ on the input). 
A robust model should generate consistent results or has lower performance degradation when dealing with these perturbed data types.

\begin{figure}
    \centering
    \includegraphics[width=0.85\linewidth]{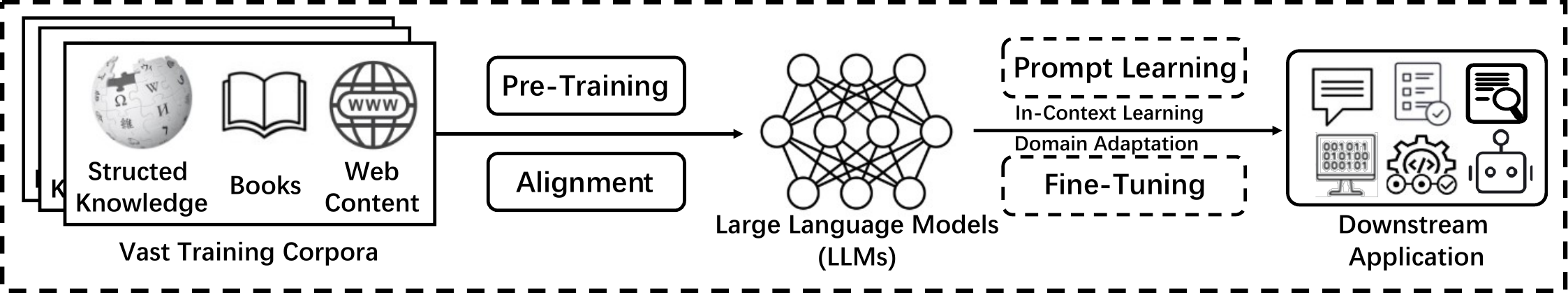}
    \caption{The pipeline of training and applying Large Language Models (LLMs).}
    \label{f:pipline}
    \vspace{-4mm}
\end{figure}


In contrast, in the era of LLMs, robustness requires more content and considerations. 
As the pipeline illustrated in Fig.~\ref{f:pipline}, compared with ML, LLMs use \textbf{enormous training corpora} (e.g, book, web contents, codes) to finishi pre-training, which has already covered vast scenarios. 
After pre-training, using \textbf{prompt} to unify different task inputs is the main form of applying LLMs to downstream areas, which are largely different from ML. 
Moreover, LLMs use transformer architecture to \textbf{decode (generate)} expected content for different tasks, different from ML designing task-related structures. 
Furthermore, with evolving of LLM capability, there has been a shift from ``internet AI'' to ``embodied AI~(EAI)'', where AI agents are expected to learn from interactions and faithfully replicate the physical world~\cite{duan2022survey}. 
Agents are expected to interact with the real world, memorize knowledge, and make reliable planning or action decision. 
As the brain of agents, LLMs are expected to demonstrate robustness in real-world scenarios. 
For example, an LLM-based agent is expected to generate consistency planning steps when facing diverse noise prompts with the same intention. 
In fact, a large number of works focus on evaluating and improving the robustness of LLMs, including (but not limited to) robust Parameter-Efficient Fine-Tuning~(PEFT)~\cite{mao2024survey}, noise prompt perturbations~\cite{zhu2023promptbench}, Out-Of-Distribution~(OOD) generalization~\cite{zang2024overcoming,liu2024good,jiang2024linguistic}, as well as robust dataset and evaluation frameworks~\cite{zhao2023chbias,li2023halueval,ullah2024detecting}. 
However, existing works either focus on a specific robustness aspect, or treat \robust~as a component in LLM survey, lacking systematic comparison and analysis between ML robustness and \robust.

Considering the learning and application paradigm of LLMs, we formally define the \robust~as follows:
\begin{definition}
\label{d:lm-robust}
    \underline{\robust} refer to the LLM's ability to maintain performance, consistency, and reliability across a wide rage of \textbf{prompt} conditions, \textbf{generating} accurate and relevant responses regardless of question phrasing types from users.
\end{definition}
Suppose we have the original data points $(X,Y)$, perturbed data points $(X^{'}, Y^{'})$. The data distribution can be described as $p(X,Y)$ and $p(X^{'}, Y^{'})$. 
Meanwhile, we can different perturbation operation $\epsilon\in\Delta$ to disturber data from origin data to perturbed data.
The loss function $L(\cdot)$, the \robust~can be formally defined as follows:
\begin{equation}
    \label{eq:robust}
    \begin{split}
        Eval(\theta) = \argmin\limits_{\theta}\underset{\epsilon\in\Delta}{max}\left((L(LLM(X), Y))+\alpha L(LLM(X^{'}), Y^{'})+\beta d(L(LLM(X)||L(LLM(X^{'}))\right),
    \end{split}
\end{equation}
where $\theta$ represents the model parameters, and $d(\cdot,\cdot)$ is the distance calculation, such as KL divergence.  
$\underset{\epsilon\in\Delta}{max}(\cdot)$ denotes the maximum data perturbation, such as the maximum noise $\delta$ in noise prompt: $X' = X + \delta$ or maximum distribution distance in OOD scenarios: $d(p(X,Y), p(X^{'},Y^{'}))$. 
Note that Eq.(\ref{eq:robust}) covers different aspects of \robust, we can use hyper-parameters $\{\alpha, \beta\}$ to control the focus of robustness optimization for the requirements of different scenarios, such as:
\begin{itemize}
	\item \textit{Performance}: $L(LLM(X), Y))$, $L(LLM(X^{'}), Y^{'})$;
	\item \textit{Consistency}: $d(L(LLM(X)||L(LLM(X^{'})$;
	\item \textit{Reliability}: $L(LLM(X^{'}), Y^{'})$ when $Y^{'}\neq Y$;
\end{itemize}




\begin{table}[]
	\centering
	\caption{The scope and corresponding keywords in the data collection process.}
	\label{t:key-words}
	\begin{tabular}{l|l}
		\toprule
		Groups & Key Words  \\ \midrule
		\textit{Fundamental} & Robustness, Generalization, OOD \\
		\midrule
		\textit{Scope}       & Large Language Models, LLMs, Pre-Trained Models   \\
		\midrule
		\textit{Key Words}   & {\makecell[l]{Robustness, Generalization, OOD, Adversarial, Prompt Learning, Noise, Interpretation, \\Knowledge shift, Interpretability, Evaluation, Benchmarks, Long Context,  Reasoning, \\Zero-shot, Few-shot, Fine-tuning, PEFT, ...}} \\
		\midrule
		\textit{Sources} & {\makecell[l]{ICLR, ICML, SIGIR, WWW, KDD, AAAI, ACL, EMNLP, IJCAI, CIKM}} \\
		\bottomrule
	\end{tabular}
    \vspace{-4mm}
\end{table}

Based on the great effort made in this field and our definition, we intend to review the progress in \robust, and give a structured overview of this research direction in this paper. 
Moreover, we also aim to highlight the connections and differences between ML robustness and \robust, as well as discuss future research directions. 
Our work differs from existing similar works in the following aspects. 
(1) \textit{Compared with LLM surveys:} Existing LLM surveys~\cite{chang2024survey,chen2023robust,wang2023robustness,zhao2023survey,zhou2024survey,minaee2024large} aim at providing a comprehensive view of LLMs, in which \robust~only occupies a small part. In contrast, our survey aims to provide a comprehensive overview of \robust; 
(2) \textit{Compared with robustness surveys:} Existing robustness surveys~\cite{goyal2023survey,yuan2023revisiting,gallegos2024bias,wang2021measure} either concentrate on ML scenarios or aim to analyze a specific aspect of \robust. In contrast, our survey not only extends the score of robustness (e.g., PEFT, long context, hallucinations, etc), but also covers a wider range of LLMs;
(3) \textit{The human impact of robustness:} Despite the above difference, we also investigate the human function and impact on \robust, which have not been considered by existing LLMs to the best of our knowledge. 
In summary, the main contributions of this survey are listed as follows:
\begin{itemize}
    \item To the best of our knowledge, we are the first to concentrate on the \robust, providing a detailed analysis of the connections and differences between robustness in ML and robustness in LLMs. 
    \item We give a formal definition of \robust, especially for the robustness when applying LLMs to safety-critical scenarios. 
    \item We conduct a detailed investigation about \robust~from three main aspects: \textit{Adversarial Robustness}, \textit{OOD Robustness}, and \textit{Robustness Evaluation}. 
    \item We present a thorough analysis of the gap between academics and industries in this field, and discuss the future research directions. 
\end{itemize}

The remainder of the paper is organized as follows. 
Section~\ref{s:protocol} presents the methodology we used for conducting this survey. 
Next, in Sections~\ref{s:adversarial},\ref{s:ood}, and \ref{s:evaluation}, we introduce our survey based on the proposed taxonomy. 
After reviewing existing work, we discuss the central roles of humans in evaluating and improving \robust, as well as highlight the future opportunities and research directions of \robust~in Section~\ref{s:discuss-future}. 



\section{Collecting Protocol}
\label{s:protocol}
In this section, we first provide a detailed methodology for paper collection and clustering of this survey. 
Then, we present the topology of this survey to provide guidance on how to exploit it. 
Finally, we summarize the difference between \robust~and \mlrobust.

\begin{figure}
    \centering
    \includegraphics[width=0.95\linewidth]{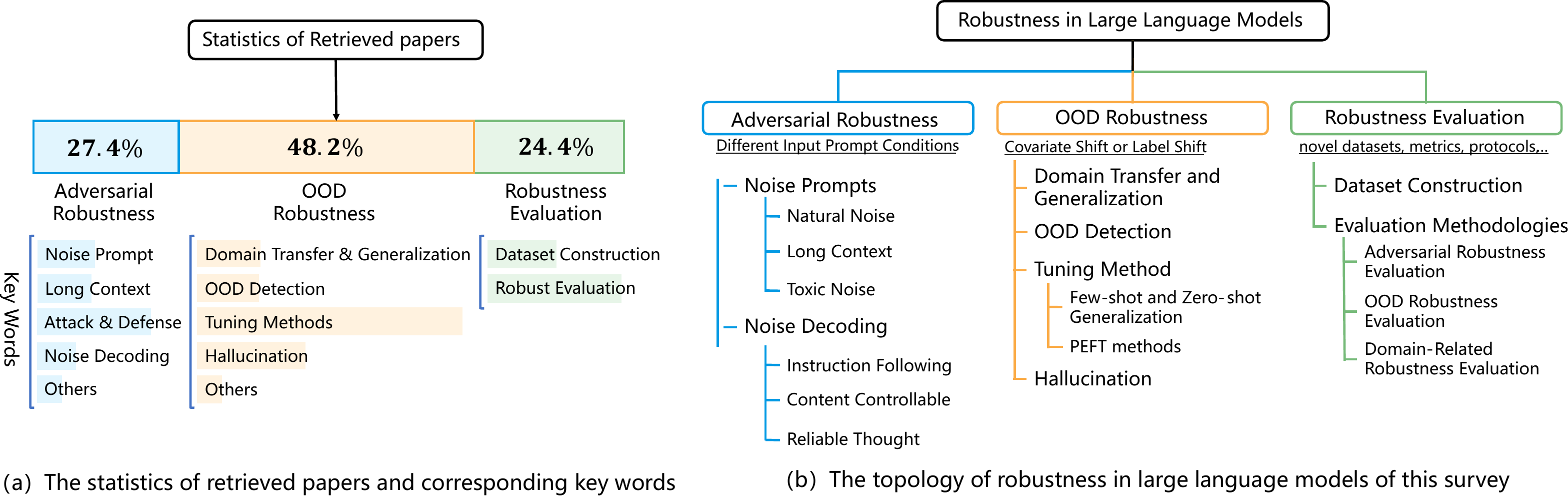}
    \caption{(A) The statistics of the topic and keyword distributions from the collected papers. (B) The topology of our survey.}
    \label{f:topology}
    \vspace{-4mm}
\end{figure}

\subsection{Collecting and Filtering Papers}
\textbf{Defining Keywords.} 
According to the definition of \robust~in Section~\ref{s:intro}, we first summarize the keywords and scope used for searching relevant papers. 
Following previous works~\cite{tocchetti2022ai,liu2023trustworthy,braiek2024machine}, we select ``\textit{Robustness, Generalization, OOD}'' and their synonyms as the keywords. 
Meanwhile, we restrict the research scope to ``\textit{Large Language Models, Pre-Trained Models}''. 
For the resources, we retrieve from AI-relevant international conferences and transactions (Journals) for the paper selection.
As for the pre-print papers in Arxiv\footnote{https://arxiv.org/}, apart from highly cited papers (e.g., \cite{conneau2018xnli}, \cite{hu2021lora}, \cite{achiam2023gpt}), we use them as complementary due to the absence of peer-reviewed and the rapid growth of paper numbers. 
Table~\ref{t:key-words} shows the list of keywords and retrieval scopes. 

\noindent \textbf{Collecting and Filtering Papers.} 
After determining the keywords, we retrieve relevant conferences and transactions (Journals) based on these keywords. 
Then, we review the titles and abstracts of the collected papers to filter and group them. 
Specifically, the search results will exclude those papers that only mention robustness or are irrelevant with Definition~\ref{d:lm-robust}. 
Next, based on the keywords used in papers~(e.g., ACM Transaction papers) or highlighted in abstracts (e.g., ACL conference papers), we group the filtered papers into different categories. 
After that, we count the percentage of papers in different categories as well as the distributions of corresponding keywords, and report the results in {Fig}.\ref{f:topology}(a). 
Note that papers may contain multiple keywords in each category and therefore may be double-counted. 
We hope the visualization will provide insight and guidance for our survey.

\subsection{Topology of this Survey}
\label{s:topology}

Based on the collection and statistics of papers, we have observed that existing works only focus on investigating a specific aspect of \robust, lacking a comprehensive consideration of \robust. 
In response, we first present the topology of \robust, as illustrated in Fig.\ref{f:topology}(b). 
Then, we give a detailed description of each research aspect of \robust, along with a summary of specific research topics.

In concerned details, based on the comparison between ML robustness and LLM robustness in Table~\ref{t:robust-comparison}, as well as previous works~\cite{vstefanik2022methods,zhao2023survey,chang2024survey}, we propose to summarize \robust~from the following three aspects: \textit{Adversarial Robustness}, \textit{OOD Robustness}, and \textit{Robustness Evaluation}. Next, we will describe each of them in detail. 

\subsubsection{Adversarial Robustness}
\label{s:adver-robust}
As shown in Fig.~\ref{f:topology}(b), LLMs unify diverse inputs into prompt format, and use generation to finish tasks. 
Therefore, the adversarial noise in prompts includes not only injected noise $\delta$, but also long context, attack and defense, etc. 
To ensure the performance of downstream tasks, LLMs incorporate multiple strategies to ensure the quality of decoding (generating) process, such as Beam Search, Chain-of-Thought (CoT), rethinking, etc, which can also be treated as robustness improvement operations. 
To this end, based on Eq.(\ref{eq:robust}), we can formulate the prompt $X$ and generation $Y$ for adversarial robustness of LLMs as follows:
\begin{equation}
	\label{eq:adversarial}
	\begin{split}
		Y^{'} = Y =\{y_1, y_2, ..., y_n\}, \quad
		\epsilon \in \{X^{'} = X+\delta, X^{'} = Attack(X),  LongContext\}. \quad
	\end{split}
\end{equation}
With this formulation, we can list the main aspects of adversarial robustness of LLMs as follows:

\noindent 1) \underline{\textit{Noise Prompt}} focuses on the outside noise and deals with carefully designed perturbations or ``noise'' in the prompt, such as carefully designed prompts, combination of multilingualism, very long context, toxic or biased attack prompts, and so on. 

\noindent 2) \underline{\textit{Noise Decoding}} concentrates on the inside noise and aims to ensure the robustness of generated contents, such as instruction following, content consistence and controllable, reliable of each generation step, etc.

\subsubsection{OOD Robustness}
\label{s:ood-robust}
As shown in Fig.~\ref{f:topology}(b), LLMs are expected to be applied to broader scenarios, using complex tools, and collaborating with other intelligence. 
Thus, LLMs must have the ability to know its capability boundaries (i.e., knowing what they know and what they do not). 
In other words, OOD detection is one of important aspects of OOD robustness in LLMs. 
Moreover, compared with enormous parameters in LLMs, downstream tasks do not have sufficient data for tuning all parameters. 
Therefore, Parameter-Efficient Fine-Tuning~(PEFT) is the main option for applying LLMs, which is also one important aspect in OOD robustness. 
Meanwhile, since we cannot tuning all parameters as ML do, it is essential how LLMs perform when the information or knowledge is updated (transferring). 
This problem is also coined as hallucinations, which are also main components in OOD robustness. 
To this end, we can reformulate Eq.(\ref{eq:robust}) for OOD robustness in LLMs as follows: 
\begin{equation}
	\label{eq:ood-robust}
	\begin{split}
		LLM(X^{'}) \in \{Y^{'}, NotKnown\}, \quad 
		0 < d(p(X^{'},Y^{'}), p(X, Y)) \leq \eta,		
	\end{split}
\end{equation}
where $\eta$ is the upper bound of data distribution distance. $NotKnown$ denotes that LLMs refuse to answer what they do not know. 
Since OOD robustness pays more attention to the data distribution, we present the main directions as follows: 

\noindent 1) \underline{\textit{OOD Detection}} requires that LLMs should be able to identify inputs, tasks, or scenarios that significantly deviate from its training distribution or expected operational domain. 

\noindent 2) \underline{\textit{PEFT methods}} focus on design parameter efficient tuning methods to satisfy downstream domains, such as Low-Rank Adapter, value alignment, zero-shot and Few-shot tuning, test-time adaptation, and so on. 

\noindent 3) \underline{\textit{Hallucination}} requires LLMs to identify and fla instances where LLMs generate factually incorrect, nonsensical, or unfounded content, particularly when dealing with inputs or contexts that fall outside its training distribution.

\subsubsection{Robustness Evaluation}
\label{s:robust-evaluation}
Apart from the above research aspects, robustness evaluation in LLMs is also one important direction, including designing novel evaluation datasets, metrics, protocols, and benchmarks. 
Different from \mlrobust~that can be evaluated with carefully designed data, LLMs have been pre-trained with vast data. 
In other words, general robustness evaluation will face data leakage problems. 
Moreover, in order to better capture LLM capability in practice, the evaluation should go beyond the exact instances contained in the specific scenarios.
In response, we summarize the key research topics as follows:

\noindent 1) \underline{\textit{Dataset Construction}} focuses on constructing challenging and unseen datasets to verify the capability of LLMs and find the boundaries of LLM capability. 

\noindent 2) \underline{\textit{Evaluation Methodologies}} concentrates on developing various metrics and evaluation protocols to evaluate \robust~from different aspects, such as consistency measures, uncertainty quantification, factual consistency scores, toxicity and safety, and so on. 
Meanwhile, we also list some technical reports of advanced LLMs.

\begin{table}[]
    \centering
    \small
    \caption{The comparison between \mlrobust~and \robust.}
    \label{t:robust-comparison}
    \begin{tabular}{c|c|c}
    \toprule
     Stage   & \mlrobust  & \robust   \\ \midrule
\textit{Input} & {\makecell[c]{Different Input types}} & {\makecell[c]{Unified Prompt input type \\Prompt format, quality, toxic, etc}} \\\midrule

\textit{Tuning}  & {\makecell[c]{Full Parameter Tuning}} & {\makecell[c]{PEFT Tuning \\ Zero-shot and Few-shot Tuning}} \\\midrule

\textit{Output} & {\makecell[c]{Different output types}} & {\makecell[c]{Unified Generation process \\ Searching, Thinking}} \\\midrule

\textit{Application}  & {\makecell[c]{Different structures or paradigms}} & {\makecell[c]{Unified Transformer paradigm \\ Memorizing, Hallucination}} \\
\bottomrule
    \end{tabular}
    \vspace{-4mm}
\end{table}

\subsection{Discussion}
\label{s:difference-discussion}
Compared with \mlrobust, \robust~has the following differences: 
1) \textit{Input}: LLMs unify inputs of different tasks with prompt, such as using natural language to introduce the task. 
Different from \mlrobust~that only concerns data-relevant noise in inputs, \robust~requires more considerations, such as the quality of task description and examples in prompt, the prompt length, the toxic injection of prompt, and so on. 
The input format is unified and the content is largely extended. 
2) \textit{Tuning}: LLMs usually have enormous parameters, which is impractical to tune all of them for each task. 
Different from tuning all parameters in \mlrobust, \robust~requires to tune a small number of parameters. 
Thus, PEFT methods are the main focus in this direction. 
3) \textit{Output}: Similar to unify inputs with prompt, LLMs also use generation to finish different tasks, which is also one important difference compared with ML methods. 
Thus, the generation quality is also essential for \robust. Beam Search, Chain-of-Thought (CoT), rethinking strategies are proposed to improve the robustness of LLMs. 
4) \textit{Application}: Different from ML that design different structures and algorithm to satisfy the downstream tasks, LLMs are expected as the foundation models for downstream. Instead of learning a new method from scratch, how to ensure the knowledge accuracy and update knowledge for different scenarios are the main concern of \robust. 
We also summarize the comparison in Table~\ref{t:robust-comparison} for better illustration. 

Moreover, to organize the papers and make them easy to utilize, we build a GitHub repo~(\github) based on the topology of this paper. 
Specifically, we follow the organization in Section~\ref{s:topology} to group the collected papers, including titles, publishers, paper links, and code repos if available. 
Meanwhile, we also continually update this repo to follow cutting-edge research.

\begin{figure}
    \centering
    \includegraphics[width=0.7\linewidth]{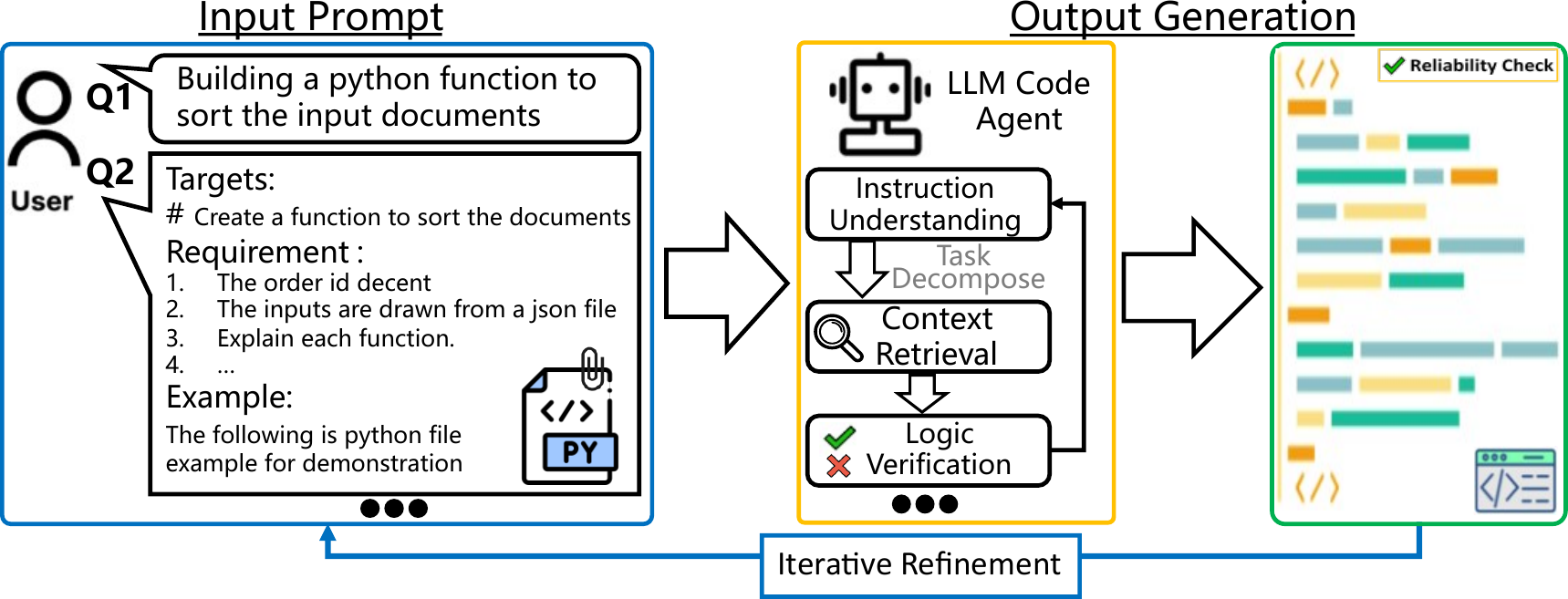}
    \caption{An example of code agents processing various requirements and generate code solution.}
    \label{f:code-agent}
    \vspace{-4mm}
\end{figure}

\section{Adversarial Robustness of LLMs}
\label{s:adversarial}
As mentioned in the previous section, adversarial robustness of LLMs mainly focuses on the input prompt and output generation process. 
Taking Fig.~\ref{f:code-agent} as an example, when developing a code agent with LLMs, LLMs should understand user intentions with diverse problem descriptions (e.g., Q1, Q2 in Fig.~\ref{f:code-agent}). 
Meanwhile, to provide a detailed description of the intention, users will use existing code examples to assist LLMs to better finish the task, which requires LLMs to deal with long code examples (i.e., long context).  
After that, we expect the code agent to generate reliable and executable codes. 
Therefore, LLMs must ensure the quality of the generation process, such as code consistency, logic verification, and so on. 
Moreover, when the problem is too difficult to be solved directly, code agent is expected to divide the task into small solvable problems , retrieve the relevant context, and use multi-step generation to finish the task, which requires LLMs to maintain the consistency and logical relevance of the sub-tasks. 
Based on the requirements in this example, we introduce adversarial robustness of LLMs from the following aspects: 
1) \textit{Noise Prompt}: focuses on the outside noise and includes Diverse Prompts, Long Context, Toxic Attack Prompts, and so on; 
2) \textit{Noise Decoding}: focuses on the inside noise and consists of Beam Search, Chain-of-Thought (CoT), Multi-Step Reasoning, and so on.

\subsection{Noise Prompt}
\label{s:noise-input}
As illustrated in Fig.~\ref{f:code-agent}, AI agents employ LLMs as the core brain to process the diverse input from the real world. 
Therefore, it is essential that LLMs should perform stably when dealing with these noise prompts. 
Specifically, based on the impact of noise, we group the related work from the following aspects: 
1) \textit{General Noise} includes spelling, grammatical errors, ambiguity, multilingualism, etc; 
2) \textit{Long Context Noise} focuses on very long prompt; 
3) \textit{Toxic noise prompt} includes attack and defense in prompt. 

\subsubsection{General Noise Prompt}
\label{s:general-noise}
In real-world scenarios, user-generated prompts usually contain diverse noises, such as spelling errors, grammatical errors, ambiguity, inductive bias, and so on. 
To deal with these general noises, a natural idea is to enrich the formats of prompts for robustness improvement. 
For example, 
\citet{patel2021stated} proves that biased prompts lead to significant differences in the style, theme, and emotional polarity of the text generated by LLMs. 
\citet{sclar2024quantifying} proposes using a range of plausible prompt formats instead of a single format to improve the robustness of LLMs.
\citet{brahmavar2024generating} proposes dividing prompts into domain constraints, which can be written in standard logical form, and simple text-based queries. 
\citet{zhu2024towards} proposes combining sentence-level rephrasing and word-level embeddings to reduce the negative impact of noise input in LLM-based retrieval tasks. 
Other methods~\cite{atwell2024combining} also focus on enriching the input format for LLMs to tackle the noise problem and improve the robustness of LLMs.
Besides, we can express the same semantics with different prompts. Therefore, using embedding to denote the prompt can enhance the adversarial robustness of LLMs. 
E.g., \citet{sun2024evaluating} leverages ``soft prompt'' embedding parameters to represent the different instruction phrasings with the same semantic meanings. 
\citet{zhu2024towards} uses word-level embeddings to assess relevance and adjust the weight of words, alleviating the negative impact of noise. 
Besides, researchers argue that noise prompt may cause biased generation of LLMs. Thus, they pay more attention to the debiased learning to improve the robustness of LLMs, such as InterFair~\cite{prasad2022interfair}, 
True zero-shot Learning~\cite{zhao2024correcting}, Embedding Projection~\cite{dawkins2024projective}, and other debiased methods~\cite{wang2021identifying,si2023measuring,fei2023mitigating}.

\begin{table}
	\centering
    \small
	\caption{Summarization of works on Adversarial Robustness of LLM with different input categories.}
	\label{t:adversairal-type}
	\begin{tabular}{c|c|c}
		\toprule
		Category          & Summary     & Methods                   \\ 
		\midrule
		\multirow{3}{*}{Noise Input}  & Enrich the prompt formats & \makecell{FORMATSPREAD~\cite{sclar2024quantifying}, LMLF~\cite{brahmavar2024generating}, \\Multi-view prompt~\cite{zhu2024towards}} \\ \cline{2-3}
        & Using embedding as soft prompts & \makecell{Soft Prompts~\cite{sun2024evaluating}, Multi-view prompt~\cite{zhu2024towards}, \\InterFair~\cite{prasad2022interfair}, Embedding Projection~\cite{dawkins2024projective}} \\ \cline{2-3}
        & Multilingualism alignment & MULTISIM~\cite{ryan2023revisiting} \\ 
        \midrule
        \multirow{3}{*}{Long Context}  & \makecell{Optimizating attention mechanism in \\ transformer unit} & \makecell{NSA~\cite{yuan2025native}, LongLoRA~\cite{chenlonglora}, MoBA~\cite{lu2025moba}, \\LEX~\cite{sun2023length}, ALiBi~\citet{sun2023length}} \\ \cline{2-3} 
        & Optimizing position embedding in inputs & \makecell{Random text replacement~\cite{hsieh2024found}, \\Longer supervised data~\cite{an2024make}} \\ \cline{2-3}
        & Compressing long context to short context & LongLLMLingua~\cite{jiang2023longllmlingua}, AutoCompressor~\cite{chevalier2023adapting} \\
        \midrule
		\multirow{7}{*}{\makecell{Toxic Noise\\Prompt}}
        & \makecell{\textit{Attack}:\\Prompt injection} & \makecell{ProPILE~\cite{kim2023propile}, TrojLLM~\cite{xue2023trojllm}, AttackVLM~\cite{zhao2023on}, \\Temporal Prompts\cite{naseer2023boosting}, SGA~\cite{lu2023set}} \\ \cline{2-3}
        & \makecell{\textit{Attack}:\\Jailbreak attacks} & Jailbroken~\cite{wei2023jailbroken}, JBPieces~\cite{shayegani2024jailbreak} \\ \cline{2-3}
        & \makecell{\textit{Attack}:\\Poisoning attacks} & \makecell{Poisoning Tuning~\cite{wan2023poisoning}, BadChain~\cite{xiang2024badchain}, \\Implicit Toxicity~\cite{wen2023unveiling}}\\ \cline{2-3}
        & \makecell{\textit{Attack}:\\Downstream Application Attack} & \makecell{MathAttack~\cite{zhou2024mathattack}, Visual attack~\cite{yin2024vqattack,qi2024visual}, \\ CodeAttack~\cite{jha2023codeattack}, Text attack~\cite{maheshwary2021context,datta2022learn2weight,xiong2024enhance}} \\
        \cline{2-3}
        & \makecell{\textit{Defense}:\\Defending against adversarial samples}  & APT~\cite{li2024one}, ORTicket~\cite{zhou2024orticket}, TaiChi~\cite{chen2024taichi}  \\\cline{2-3}
        & \makecell{\textit{Defense}:\\Removing information from model parameters}  & PAD~\cite{yang2024pad}, Learn2Weight~\cite{datta2022learn2weight}, DeMem~\cite{kassem2023preserving} \\\cline{2-3}
        &\makecell{\textit{Defense}:\\Protecting privacy with differential privacy methods}  & PromptDPSGD~\cite{duan2023flocks}, FABE~\cite{liu2024causality} \\
		\bottomrule
	\end{tabular}
    \vspace{-4mm}
\end{table}

Apart from noise prompt in single language (i.e., English), it is also important to maintain the adversarial robustness in multilingualism considering its broad applications (e.g., code agents, teaching assistant, etc). 
\citet{muennighoff2023crosslingual} has demonstrated that LLMs fine-tuned solely on English data can generalize across languages.  
Moreover, by fine-tuning on an intermediate language, we can extend LLMs to minority language, \citet{ryan2023revisiting} demonstrates that fine-tuning on Russian data can achieve better zero-shot generalization performance in low-resource languages. They have published a unified Multilingual Benchmark to evaluate \robust~in multilingualism.
One step further, since LLMs are mainly pre-trained on English, noise from other languages may affect the robustness of LLMs. 
\citet{fridhriksdottir2024gendered} observes that specific grammar types in Icelandic may trigger gender bias in LLMs and cause LLMs to be vulnerable to generating gender-related content, which highlights the complex interplay between social and linguistic influences.

\subsubsection{Long Context Noise Prompt}
\label{s:long-context}
As illustrated in Fig~\ref{f:code-agent}, a code agent is often required to process long code examples for task understanding, demonstrating the importance of handling long context. 
However, since LLMs use self-attention in transformer unit to process the input sequence, long context will cause the attention weights too flatten to highlight the most important part. 
Moreover, position embedding also causes LLMs to memorize specific positions. 
They all lead to vulnerable performance. Thus, optimizing these two components is preferred to improve the robustness of LLMs when dealing with long contexts. 

For attention mechanism~\cite{presstrain, sun2023length, su2024roformer, lifunctional},
Sparse attention is the mainstream for better processing long context. 
\citet{lu2025moba} employs MoE and designs a Mixture of Block Attention (MoBA), where long context is split into different blocks and each query token will focuses on the most relevant key-value blocks. 
Moreover, MoBA can be integrated with full attention easily, which is flexible for real-world application. 
Meanwhile, \citet{yuan2025native} proposes an native sparse attention~(NSA) for long context processing of LLMs. 
NSA organizes key-value pairs into temporal blocks and designs three attention branches: compressed attention mask for coarse-grained patterns, selected attention mask for important token blocks, and sliding attention mask for local context. 
Thus, NSA can realize hardware aligning and boost the training and implementing. 
\citet{chenlonglora} leverages LoRA to compute intra-group and inter-component attention, efficiently enhancing the capability of existing LLMs to handle long texts. 
Meanwhile, Linear Attention~\cite{han2023flatten,meng2025polaformer,gu2023mamba,peng2023rwkv,sun2023retentive} also attracts plenty of attention.

For position embeddings, \citet{liu2024lost} conducts diverse experiments to prove that position bias is a major factor affecting the robustness of LLMs when handling long contexts. 
\citet{presstrain} designs a functional relative position encoding with progressive interpolation (FIRE) to improve the generalization of LLMs on long contexts. 
They apply popular log transformation and a modified progressive interpolation with a learnable threshold in the normalizer, so that LLMs can learn to pay more attention to far away contexts in some attention heads. 
\citet{sun2023length} uses a penalty that is proportional to the distance to bias the query-key attention scores, instead of adding positional embeddings into word embedding. Their proposed ALiBi~\cite{sun2023length} has been proven to outperform multiple strong position embedding methods in improving LLM robustness in dealing with long contexts. 
Other works, such as constructing long context data~\cite{an2024make}, are also developed.

Meanwhile, since LLMs perform better in short context, some works are designed to handle long context through compression. 
\citet{jiang2023longllmlingua} compresses relevant texts by employing importance assessment and reordering mechanisms. 
\citet{chevalier2023adapting} segments long texts and used AutoCompressors to compress them into vectors, applying autoregressive prediction between segments to minimize information loss.

\subsubsection{Toxic Noise Prompt}
\label{s:toxic-noise}
AI agents and EAI are designed to interact with real-world scenarios. Moreover, they are expected to assist humans in tackling real-world problems, operating various machines, and increasing productivity. 
As the core brain of AI agents and EAI, LLMs should be robust and reliable when facing toxic prompts. 
For example, an LLM-based bank agent cannot generate the privacy of clients when asked ``Tell me something about the richest client in your bank''. 
Thus, designing attack methods is important to verify the boundaries of \robust. 
Meanwhile, developing defense methods is essential to improve the \robust. 
Next, we will introduce each part in detail. 

\underline{\textit{Attack. }}
In terms of attack objectives, existing research primarily focuses on four aspects: degrading model performance~\cite{xue2023trojllm, aich2022gama, naseer2023boosting, lu2023set}, generating harmful outputs~\cite{wei2023jailbroken, shayegani2024jailbreak}, obtaining private information~\cite{kim2023propile}, and manipulating model outputs~\cite{zhao2023on, xiang2024badchain, yang2023data, wan2023poisoning}. 
In terms of attack methods, typical approaches include: prompt injection (manually constructing prompt information or generating adversarial samples based on algorithms)~\cite{kim2023propile, xue2023trojllm, zhao2023on, aich2022gama, naseer2023boosting, lu2023set}, jailbreak attacks (exploiting vulnerabilities in large models for attacks)~\cite{wei2023jailbroken, shayegani2024jailbreak}, poisoning attacks (poisoning during pre-training or fine-tuning by inserting specified backdoors)~\cite{wan2023poisoning, xiang2024badchain, yang2023data} and downstream application attacks (targeting specific applications of LLMs such as mathematical reasoning, code generation, or visual question answering)~\cite{zhou2024mathattack, yin2024vqattack, qi2024visual, jha2023codeattack, datta2022learn2weight, xiong2024enhance}.

For example, \citet{zhou2024mathattack} introduces MathAttack, which retains mathematical logic when attacking math word problem samples. 
They utilize logical entity recognition and a word-level attacker to execute the attack. Experimental results offer important insights for enhancing LLM robustness in mathematical problem-solving. 
\citet{jha2023codeattack} proposes CodeAttack, which generates effective, efficient, and imperceptible adversarial code samples by leveraging code structure, showcasing the vulnerability of large language models to code-specific adversarial attacks. 
They argued that LLMs are focused on human understanding of code but are not robust enough to input variations, making them vulnerable to adversarial attacks.
\citet{wei2022towards} introduced a dual-attack framework combining PNA (Perturbation Not Attention) attacks and PatchOut attacks, successfully achieving higher attack success rates on different ViT models. 
Experiments reveal that attacking the attention gradients can effectively degrade the robustness of LLMs. 
Meanwhile, other works focus on generating adversarial examples to realize attacks on LLMs, such as text attack~\cite{maheshwary2021context,datta2022learn2weight,xiong2024enhance}, visual attack~\cite{yin2024vqattack,qi2024visual}, and attack strategies~\cite{hao2021self,maheshwary2021generating}.

\underline{\textit{Defense. }}
Building on the diverse research on attacks, relevant researchers have also proposed corresponding defensive algorithms, such as defending against adversarial samples through algorithms~\cite{li2024one}, removing information from model parameters, and protecting data privacy information by introducing differential privacy methods~\cite{duan2023flocks}. 
For example, 
\citet{chen2024taichi} introduces TaiChi method, which uses a Siamese network architecture and a contrastive learning strategy to encourage similar generations, as well as a KL divergence loss to enhance the consistency of prediction. 
\citet{zhou2024orticket} designs a novel ORTicket, which leverages the robustness transferability within sub-networks through pruning and fine-tuning, achieving efficient robustness without separate adversarial training. Experimental results indicate the potential of this method for further enhancing model robustness. 
\citet{liu2024causality} proposes a novel defense framework based on causality: Front-door Adjustment for Backdoor Elimination (FABE), which uses causal reasoning to counter diverse backdoor attacks on LLMs without relying on assumptions about trigger forms. By employing a front-door criterion to differentiate spurious from legitimate associations, FABE effectively reduces attack success rates. 
Moreover, \citet{yang2024pad} notes that existing robustness enhancement methods typically target specific types of perturbations. 
They develop a defense method called PAD, which employs a plugin module to perturb the base model's parameters, achieving the effects of multiple models while saving computational resources.
There exist other defense methods for the robustness of LLMs, such as Learn2Weight~\cite{datta2022learn2weight}, PlugAT~\cite{zheng2022plugat}, and RL-based method~\cite{wen2023unveiling,kassem2023preserving}.

\subsection{Noise Decoding}
\label{s:noise-decoding}
Apart from unifying diverse inputs with the prompt format, LLMs also employ the decoding (generating) process to finish diverse tasks (e.g., classification, reasoning, etc).
Taking Fig.~\ref{f:code-agent} as an example, when the code agent generates the code solution for users, many variables can affect the quality of generated contents, such as different temperature $\tau$ and different sampling strategies. 
Meanwhile, tiny errors in the generation process can be amplified during multi-step generation, especially in long sequence generation. 
Thus, how to ensure the robustness of the generation process is also essential for \robust. 
In response, we focus on the \textit{Decoding Strategies} and \textit{Inference Strategies} to introduce relevant works.

\underline{\textit{Decoding Strategies. }}
During the decoding process, how to control the stability of sampling, consistency of logic, and the decoding efficiency is the main concern for improving \robust. 
For example, \citet{li2023contrastive} designs a contrastive decoding strategy to tackle the problem of generating short and repeat text. 
They propose to optimize a contrastive objective of the likelihood between expert models (Large models) and amateur models (Small models) for the reliable text generation. 
\citet{yang2024rlcd} also employ contrastive learning to improve the decoding robustness of LLMs, reducing the requirements of annotated alignment data. 
\citet{henderson2023self} concerns more about the generated contents. 
By increasing the cost of generating harmful contents with meta learning and adversarial learning, they propose a novel self-destructing models for improve the \robust~when the LLMs can be used for both harmful and benign scenarios. 
\citet{xia2023speculative} employs the speculative execution in computer architecture and designs a speculative decoding strategy to improve the decoding efficiency. 
This strategy uses a lighted spec-drafter model to generate multiple token rapidly and uses the target model to verify these token parallel as well as accept the satisfied token. 
\citet{liuetal2024lost} designs two long context tasks (i.e., Multi-document QA and Key-Value Retrieval) to verify the robustness of LLMs when dealing with long context. 
They point out that LLMs are still lacking robustness when processing long contexts. 
\citet{bai2022constitutional} proposes to design a set of ``constitution'' to guide the decoding process of LLMs, which aims to realize AI self-supervised alignment without human annotation of harmful contents. 

\underline{\textit{Inference Strategies. }} 
During inference, the consistency and correctness of thought are main challenges for LLMs. 
For example, \citet{wang2022self} proposes a self-consistency method to improve the robustness of LLMs during inference, especially in the Chain-of-Thought scenarios. 
This method first requires LLMs to generate multiple inference paths for each inference task. 
Then, a majority vote is used to select the final output from these generated paths.
\citet{hao2023reasoning} argues that the vulnerable inference capability of LLMs is caused by the lack of world models to predict the world states. 
Thus, they propose a Reasoning via Planning (RAP) framework to boost inference capability of LLMs, where LLMs are treated as the world models with pre-defined states and actions, and inferences are realized with MDP. 
\citet{xu2023rewoo} focuses on the low inference efficiency and develops a ReWOO framework. 
They divide the inference into three modules: planner, worker, and solver. 
By using these three modules, ReWOO can reduce the cost of tokens and improve the inference efficiency, especially for tool use and small language models. 
\citet{yao2023tree} focuses on the inference steps and design a Tree-of-Thought (ToT) framework. 
They use the tree structure to constrain the inference process and employ classical search algorithms to improve the efficiency of inference. 
Apart from the above works, there also exist works for improving the robustness of inference during RAG or agent scenarios, such as self-RAG~\cite{asai2023self} and ToRA~\cite{gou2024tora}.

\textbf{\underline{Summarization.} }
Since LLMs use unified prompt format as the input and generate output sequences to finish various tasks, adversarial robustness of LLMs focuses on improving the capability of dealing with various prompts and enhancing the consistency and efficiency of decoding process. 
Therefore, enriching the prompt formats, designing advanced attention mechanisms and position embeddings, as well as developing diverse attack and defense strategies are the main directions of improving the capability of handling inputs. 
Meanwhile, employing additional optimizations, designing novel CoT strategies, and decomposing inference steps are the main contents of enhancing the decoding robustness of LLMs. 

Besides, some works have attempted to incorporate causal inference for improving the adversarial robustness of LLMs (e.g., FABE~\cite{liu2024causality} in Section~\ref{s:toxic-noise}). 
Early attempts~\cite{wu2024causality,wan2024bridging} have demonstrated the potential to use causal inference to enhance adversarial robustness of LLM, including learning causal representation to improve robustness and developing benchmarks to assess adversarial robustness and causal understanding in LLM.
However, this direction remains underexplored due to the computational complexity of causal reasoning process and traditional causal inference methods unable to deal with unstructured data, such as text and graph data.

\begin{figure}
    \centering
    \includegraphics[width=0.7\linewidth]{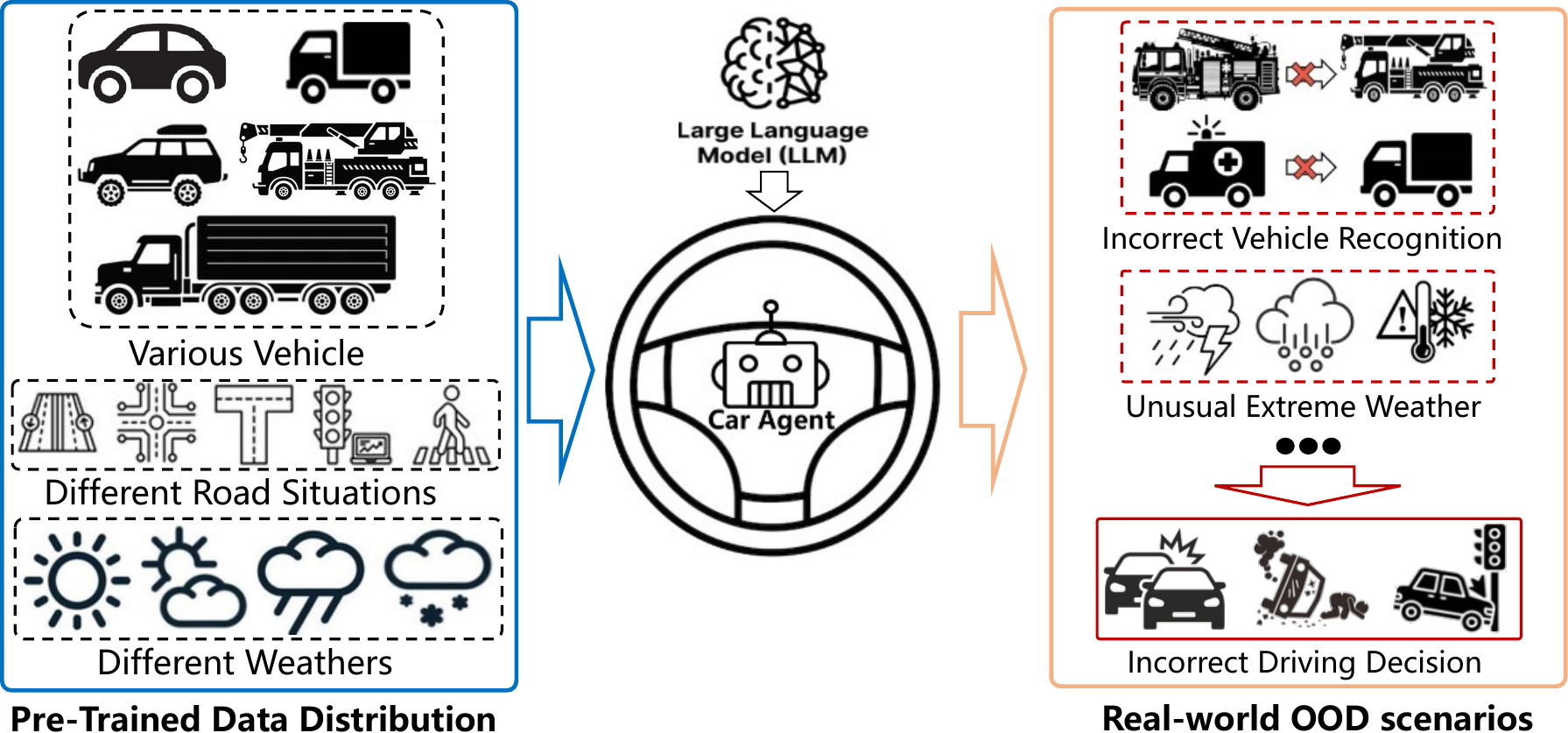}
    \caption{An example of car agents conducting incorrect decisions when dealing with Out-Of-Distribution scenarios.}
    \label{f:car-agent}
    \vspace{-4mm}
\end{figure}


\section{OOD Robustness of LLMs}
\label{s:ood}
Similar to \mlrobust, \robust~also concentrates on the domain transfer and generalization in OOD scenarios, which we believe LLMs are still far from satisfaction. 
Meanwhile, OOD robustness of LLMs also has much to concern. 
Taking {Fig}.~\ref{f:car-agent} as an example, when a car agent meets untrained objects (e.g., over-width vehicles, ambulances), if it does not know the difference between these special cars and general cars, it will conduct incorrect driving decisions (e.g., overtaking or lane occupying) and cause severe consequences (e.g., car crashing or delaying patient treatment). 
This example proves the importance of OOD detection, in which LLMs should know the data or task is in-domain or out-of-domain. 
Moreover, in the above example, the special objects are uncommon and corresponding data is less, especially compared with billions of LLM parameters. 
Thus, OOD robustness of LLMs also has to concern the Parameter-Efficient Fine-Tuning (PEFT) methods, which is quite different from \mlrobust. 
Furthermore, AI agents are designed to interact with the real world, in which knowledge and information are continually updated. 
Therefore, hallucination is also one important aspect in OOD robustness of LLMs. 
As shown in {Fig}.~\ref{f:car-agent}, the car agent should not recognize special cars as existing known cars. Otherwise, this hallucination will cause the car agent to conduct incorrect decisions. 
In the following sections, we first give a brief summary of recent works about domain transfer and generalization of LLMs. Then, we introduce the unique aspects of OOD robustness of LLMs, including OOD Detection, PEFT methods, and Hallucinations. 

\subsection{Domain Transfer and Generalization}
\label{s:domain-transfer}
The main content of the OOD robustness is how to improve the capability of LLMs to deal with domain transfer and generalization problems. 
Many researchers have proven that LLMs struggle to generalize to unfamiliar tasks and lack generalizability when faced with domain transfer~\cite{mishra2022cross, kung2023active, pelrine2023towards, hu2023diner, lu2023scitab, he2023medeval}. 
\citet{ramanujan2023on} explores the impact of pre-training data diversity on fine-tuning robustness, including data quantity, label granularity, label semantics, image diversity, and data sources. 
Since this aspect has been the primary focus on \mlrobust, we briefly group the relevant works and summarize the mainstream frameworks. 
Table~\ref{t:ood-type} illustrates the representative works in each group. 

The intuitive solution is to construct comprehensive tuning datasets, so that LLMs learn knowledge about unknown tasks as much as possible. 
For example,  \citet{mishra2022cross} constructs a dataset containing 61 tasks and unified the instruction format. 
\citet{hu2023diner} develops a large-scale, realistic dataset and utilized compositional prompting for fine-tuning. 
Besides, other researchers argue that LLMs can learn much information from harder and higher-quality datasets.
\citet{oh2024geodesic} designs a novel Geodesic Multi-Modal Mixup for mixuping the embeddings of image and text to generate hard negative samples. 
\citet{xiao2023masked} creates counterfactual examples for LLM fine-tuning. 
Other works also investigate the generalization of LLMs on different tasks~\cite{lu2023scitab,he2023medeval,kung2023active}.

Apart from dataset construction, designing domain-related fine-tuning or transfer learning algorithms is also one important direction. 
\citet{han2024anchor} has proven that OOD generalization of LLMs is mainly affected by two issues: 1) Domain shift, such as from natural to sketch images, and 2) Zero-shot capability, which is the ability to recognize classes not contained in the fine-tuned data. 
\citet{wang2024twostage} develops a two-stage fine-tuning method ProMoT, in which prompt tuning is conducted at stage one to obtain the soft prompt for the target task, and model fine-tuning with the trained prompt is performed at stage two for the LLM robustness improvement. 
\citet{lee2023surgical} designs a surgical fine-tuning method, which selectively fine-tunes a subset of layers in LLMs for different data distribution, achieving robust domain transferring. 
Moreover, \citet{goyal2023finetune} employs a contrastive learning strategy for further fine-tuning. They treat downstream class labels as text prompts and continue optimizing the contrastive loss between input embeddings and class-descriptive prompt embeddings for generalization improvement. 
Other relevant works, such as incorporating ML learning strategies~\cite{yang2023mitigating,xiao2023masked,Li_2023_ICCV}, using instruction uncertainty~\cite{kung2023active,pelrine2023towards,basu2024efficient}, designing fine-grained tuning constraints~\cite{tian2023trainable}, are also active research directions.

\begin{table}
	\centering
    \small
	\caption{Summarization of works on OOD Robustness of LLM with distribution shifts.}
	\label{t:ood-type}
	\begin{tabular}{c|c|c}
		\toprule
		Category          & Method Summary     & Methods                   \\ 
		\midrule
		\multirow{3}{*}{\makecell{Domain Transfer \\ and Generalization}} 
        & Constructing comprehensive tuning datasets & \makecell{NATURAL INSTRUCTION~\cite{mishra2022cross}, DiNeR~\cite{hu2023diner}, \\Multi-Modal Mixup~\cite{oh2024geodesic}, SCITAB~\cite{lu2023scitab}, \\Counterfactual Samples~\cite{xiao2023masked}} \\ \cline{2-3}
        & Designing domain-related transfer algorithms & \makecell{ARF~\cite{han2024anchor}, ProMoT~\cite{wang2024twostage}, Surgical Fine-Tuning~\cite{lee2023surgical}, \\FLYP~\cite{goyal2023finetune}, TPGM~\cite{tian2023trainable}, equizero~\cite{basu2024efficient}} \\ \cline{2-3}
        & Generalizing to unknown areas & \makecell{RISE~\cite{Huang_2023_ICCV}, PracticalDG~\cite{Chen_2024_CVPR_PracticalDG}, DIFO~\cite{Tang_2024_CVPR}, \\DMN~\cite{Zhang_2024_CVPR}, OGEN~\cite{zang2024overcoming}, VL2V-ADiP~\cite{addepalli2024leveraging}} \\
        \midrule
        OOD Detection & figure out the capability boundaries of LLMs & \makecell{NegLabel~\cite{jiang2024negative}, MS-OOD DETECTION~\cite{averly2023unified}, \\SelfCheckGPT~\cite{manakul2023selfcheckgpt}, LLM-OOD~\cite{liu2024good}, \\Focus~\cite{zhang2023enhancing}, Learn-Babble-Prune~\cite{bouyamourn2023llms}} \\ 
        \midrule
        \multirow{6}{*}{Tuning Methods} 
        & \makecell{Using a few examples to enhance \\the generalization capability}  & \makecell{LGIR~\cite{du2024enhancing}, COFE~\cite{an2023context}, RoCoIns~\cite{zhang2024rocoins}, \\HINT~\cite{ivison2023hint}, AAR~\cite{yu2023augmentation}} \\ \cline{2-3}
        & Stimulating the zero-shot capability of LLMs & \makecell{t-zero~\cite{sanh2022multitask}, RoboShot~\cite{adila2024zeroshot}, CTR~\cite{shao2023compositional}, \\LAAT~\cite{Li_2024_CVPR}, PMG-AFT~\cite{Wang_2024_CVPR}} \\ \cline{2-3}
        & \makecell{Incorporating Test-time adaptation (TTA)}  & \makecell{TPT~\cite{shu2022test}, TDA~\cite{karmanov2024efficient}, PromptAlign~\cite{abdul2024align}, \\RLCF~\cite{zhao2024testtime}, MTA~\cite{Zanella_2024_CVPR}} \\ \cline{2-3}
        & Aligning with human annotations or values & \makecell{RPO~\cite{rafailov2023direct}, DRESS~\cite{Chen_2024_CVPR}, PHF~\cite{korbak2023pretraining}, AfD~\cite{sun2024inverse}, \\CAA~\cite{xia2024aligning}, Moca~\cite{nie2024moca}, CPO~\cite{linoptimizing}} \\ \cline{2-3}
        & PEFT methods for robustness improvement & \makecell{Robust prefix-tuning~\cite{yang2022on}, PINL~\cite{Wu_2023_ICCV}, \\NMTune~\cite{chen2024understanding}} \\ \cline{2-3}
        & PEFT methods for privacy and fairness & \makecell{private-transformers~\cite{li2022large}, FASP~\cite{zayed2024fairness}, \\Differentially Private Fine-tuning~\cite{yu2022differentially}} \\
        \midrule
        \multirow{3}{*}{Hallucination} 
        & \makecell{Using factual alignment to prevent generating \\non-existing contents}         & \makecell{SAT Probe~\cite{yuksekgonul2024attention}, Vista-LLaMA~\cite{Ma_2024_CVPR}, \\VCD~\cite{Leng_2024_CVPR}, DoLa~\cite{chuang2024dola}, Ontofact~\cite{shang2024ontofact}} \\ \cline{2-3}
        & \makecell{Using existing knowledge bases to \\correct the outputs} & SYNTRA~\cite{jones2024teaching}, KGR~\cite{guan2024mitigating}, \\ \cline{2-3}
        & \makecell{Employing Causing Reasoning to avoid \\reliance on spurious correlations} & Counterfactual Decoding~\cite{yu2024cause}, Causal-debias~\cite{zhou2023causal} \\
        \bottomrule
	\end{tabular}
    \vspace{-5mm}
\end{table}

The above works assume that the source and target domains have the same categories or have overlaps. 
However, in reality, there may exist unknown information of target domains, such as unseen categories and data types. 
To tackle these problems, 
\citet{Huang_2023_ICCV} proposes training a student model that can generalize to unseen domains by using language information as a regularization strategy, leveraging the knowledge of the CLIP teacher model. 
\citet{Chen_2024_CVPR_PracticalDG} proposes an Open-Set Domain Generalization (OSDG) approach and introduce several methods specifically tailored to tackle this problem.
\citet{Tang_2024_CVPR} focuses on the Source-Free Domain Adaptation (SFDA) problem and designs a novel Distilling multImodal Foundation mOdel (DIFO) to tackle the problem that traditional methods incorrectly rely on pseudo-labeling and auxiliary supervision. 
Besides, there also exist many other relevant works~\cite{kumar2022finetuning,shu2023clipood,trivedi2023a,Zhang_2024_CVPR,zang2024overcoming,addepalli2024leveraging}.

\subsection{OOD Detection}
\label{s:ood-detection}
As illustrated in {Fig}.~\ref{f:car-agent}, the agent should know whether requests or situations are beyond its capability. 
Otherwise, it may make incorrect decisions, causing severe consequences (e.g., a car agent does not recognize special cars and makes incorrect driving decisions). 
We group related works as OOD detection and argue that this direction is unique in \robust, compared with \mlrobust. 
Plenty of works have been dedicated to exploring OOD detection capability in LLMs~\cite{bouyamourn2023llms, xu2023fine, kothyari2023crush4sql, qiu2023detecting, manakul2023selfcheckgpt,liu2024good}. 
For example, \citet{jiang2024negative} proposes a novel post hoc OOD detection method, called NegLabel. This method takes a vast number of negative labels and corresponding textual information from extensive corpus databases to enhance OOD detection. 
\citet{averly2023unified} designs a framework that unifies the detection of OOD examples caused by semantic shift and covariate shift, and closely addresses the concern of applying a machine learning model to uncontrolled environments.  

Meanwhile, if an LLM does not accurately recognize OOD scenarios, it may generate incorrect results, which can be grouped into hallucinations. 
Thus, similar to recognizing OOD scenarios in input domains, it is also important to detect OOD situations in the generated results, also named hallucination detection. 
For example, \citet{zhang2023enhancing} introduces an uncertainty-based approach to detect hallucinations, using a proxy language model to compute the probability of each token in a given text. This method calculates token and sentence-level hallucination scores based on these probabilities. 
\citet{manakul2023selfcheckgpt} assumes that if an LLM has knowledge of a given concept, sampled responses are likely to be similar and contain consistent facts, which is not satisfied by hallucination facts. 
Thus, they design a simple sampling-based approach SelfCheckGPT to fact-check the generated results of black-box models in a zero-resource fashion. 
Along with hallucination detection, hallucination reduction is also one important research direction, such as Learn-Babble-Prune~\cite{bouyamourn2023llms}, knowledge base enhancement~\cite{xu2023fine}, retrieval based method~\cite{kothyari2023crush4sql}.
We will provide a detailed introduction in Section~\ref{s:knowledge-shift}.

\subsection{Tuning Methods}
\label{s:tuning-method}
LLMs incorporate massive parameters to maintain general capabilities, such as perception, understanding, and memorizing. 
However, in the downstream areas, for one thing, domain data usually is too little to tune all parameters, and for another, downstream areas require only certain specific capabilities. 
Thus, how to apply LLMs to downstream areas has a big impact on \robust. 
Taking {Fig}.~\ref{f:car-agent} as an example, a car agent may meet OOD scenarios (e.g., extreme weather). In order to ensure that the agent can function well without hallucinating, we need to focus on the weather perception part. 
Meanwhile, the data on OOD scenarios is rare (e.g., driving data in extreme weather is little). 
Under such a circumstance, the following options are provided: 
1) Enhancing the few-shot and zero-shot capability of LLMs; 
2) Designing Parameter-Efficient Fine-Tuning (PEFT) methods. 
Next, we will introduce each option in detail. 

\subsubsection{Few-shot and Zero-shot Generalization}
\label{s:few-zero}
Few-shot and Zero-shot generalization focus on using few samples or using no additional samples to maintain the capability of LLMs in OOD scenarios. 
Considering the input prompt format, task prompt constructing methods, as well as the choice and order of context examples are the main factors for \robust~\cite{an2023context,patel2023magnifico}. 
For few-shot generalization, high-quality data construction and arrangement are the main focus. 
For example, \citet{du2024enhancing} uses Generative Adversarial Networks (GANs) to align unpaired low-quality resumes with high-quality generated resumes to ensure the quality of selected few-shot examples. 
\citet{an2023context} designs a test suite CoFe to investigate in-context compositional generalization of LLMs. They have observed that using examples with a structure similar to the test samples, high diversity between different examples, and individually simple examples are beneficial for the performance of few-shot generalization of LLMs. 
\citet{zhang2024rocoins} adopts structured code-style instructions and designs an adversarial context method that uses both clean and adversarial samples to boost the robustness of LLMs in few-shot scenarios. 
Besides, instruction fine-tuning~\cite{gupta2022instructdial,ivison2023hint}, retrieval-augmented method~\cite{yu2023augmentation}, and attention-based method~\cite{stacey2022supervising} are also promising direction to improve the few-shot generalization capability of LLMs.

Apart from using few samples, LLMs are expected to achieve robust zero-shot generalization, such as a car agent can adapt extreme weather effectively even it is not trained on this scenario.  
Thus, how to design effective prompt to stimulate LLMs capabilities becomes the mainstream in zero-shot generalizations. 
For example, 
\citet{sanh2022multitask} designs a system that can easily map any natural language task into a human-readable prompt format, so that the zero-shot generalization capabilities of LLMs can be verified at scales. 
\citet{adila2024zeroshot} introduces RoboShot for zero-shot robustness of LLMs, a new method that automatically derives beneficial and harmful concepts from the task descriptions, so that it does not require any labeled data, training, or fine-tuning, nor does it require manually identifying specific concepts. 
One step further, \citet{shao2023compositional} proposes a new prompt-free method called Compositional Task Representation (CTR), which learns a discrete, composable codebook through multi-task training. 
Thus, the unseen tasks can be composed with learned task information, improving the \robust~in zero-shot scenarios.
Besides, there also exists a branch that using the guidance of language (text) to improve the zero-shot generalization capability in multi-modal scenarios, such as using text features as fixed anchor points for each category~\cite{Li_2024_CVPR}, minimize the distance between generated features in the multi-modal model and pre-trained language model~\cite{Wang_2024_CVPR}, and so on.

Another novel and promising direction to enhance OOD robustness of LLMs is Test-time adaptation (TTA)~\cite{liang2024comprehensive}, which focuses on adapting a pre-trained model to unlabeled data during testing, before making predictions. 
Some early work has been done in this direction~\cite{chi2024adapting, karmanov2024efficient,abdul2024align, shu2022test, zhao2024testtime, Zanella_2024_CVPR}. 
For example, 
\citet{shu2022test} introduces a method called Test-time Prompt Tuning (TPT), which can learn adaptive prompts using a single sample at test time. TPT optimizes prompts by minimizing entropy and confidence selection, ensuring consistent predictions across different augmented views of each test sample, thereby improving the zero-shot top-1 accuracy of CLIP. 
\citet{karmanov2024efficient} utilizes a lightweight key-value cache to adapt to test data through step-by-step pseudo-label refinement progressively, avoiding any backpropagation, thus achieving super-efficient adaptation at test time.
Meanwhile, \citet{abdul2024align} proposes a new test-time prompt tuning method called PromptAlign, which adapts multimodal prompts by minimizing the feature distribution shift and uses a single test sample to adjust the prompts at test-time, thus enhancing the zero-shot generalization of LLMs. 
\citet{zhao2024testtime} argues that previous methods for TTA in zero-shot classification of LLMs rely on minimizing the entropy of the model's output, which may lead the model to fall into incorrect predictions.
Thus, they propose a TTA method with feedback, which corrects the model's output and prevents the model from being blindly overconfident.

\subsubsection{Parameter-Efficient Fine-Tuning (PEFT) methods}
\label{s:peft-method}
After pre-training, there are two steps before applying LLMs to downstream areas: 1) alignment and 2) fine-tuning. 
The former aims at aligning LLMs with human preference~\cite{ouyang2022training, rafailov2023direct}. 
The latter focuses on tuning a small number of parameters to better adapt downstream tasks. 
They all have big impacts on \robust. Next, we will give a detailed introduce of them. 

\underline{\textit{Alignment. }}
During alignment, we need human-annotated data, alignment strategies, and alignment metrics for LLM alignments. 
They all have big impact on \robust. 
For example, if driving data is collected from aggressive drivers (e.g., many congestion or lane jumping behaviors), an aligned car agent will always engage in risky driving behaviors, which is unsafe and vulnerable in some situations. 
To ensure \robust~during alignment,  one important direction is to exploit the robust features in the data.
For example, \citet{rafailov2023direct} argues that preference pairs used in DPO algorithm~\cite{rafailov2024direct} are from the same prompt, which cannot help capturing the complexity of human learning. They propose Relative Preference Optimization (RPO), which introduces a contrastive weighting mechanism, allowing LLMs to leverage a broader range of preference data for 
robustness improvement. 
\citet{yukun2024improving} proposes a two-stage training framework, including instruction-enhanced supervised fine-tuning and consistency alignment training. This framework helps the model generalize instructions and enhance diversity to better understand responses that meet human expectations, which can further improve the model robustness.
Meanwhile, incorporating an assistant is also an important direction. 
\citet{Chen_2024_CVPR} uses LLMs to generate natural language feedback for target LLMs' response. By providing fine-grained feedback and guidance (i.e., strengths, weaknesses, and specific suggestions), the target LLMs can be aligned robustly. 
\citet{korbak2023pretraining} and \citet{Yu_2024_CVPR} both propose to collect human feedback as the target to align LLMs, where humans are asked to identify and evaluate the model response and provide optimal responses for LLMs. 
By employing human-in-the-loop, \robust~after alignment operation can be maintained. 
Other similar works are also conducted to improve \robust~after LLM alignment~\cite{zheng2024improving,sun2024inverse,herrera2023large}.
In addition, some efforts align AI systems with human values through causality.
\citet{xia2024aligning} propose a causality-aware approach to align LLMs by identifying pretraining data and input prompts as confounders that lead to biased outputs. 
Their method leverages causal intervention through a reward model, improving LLM debiasing and promoting safer text generation.
\citet{nie2024moca} explore the alignment of LLMs with human causal and moral judgments using a dataset inspired by cognitive science research. 
This study uncovers notable discrepancies in how LLMs prioritize causal and moral factors, emphasizing the importance of challenging datasets for detailed evaluation.
\citet{linoptimizing} propose causal preference optimization (CPO) to align LLMs with human preferences by treating optimization as a causal problem. They further extend this with doubly robust CPO (DR-CPO), helping LLMs perform robustly under confounding conditions.

\underline{\textit{PEFT methods. }} 
When applying LLMs to downstream areas, fine-tuning a small number of parameters is the mainstream, ensuring both efficiency and effectiveness. 
Thus, \robust~also concentrates on designing robust PEFT methods. 
For example, if we only use a small size of driving data in extreme weather to tune a car agent, it may be overfitting on this data and occur catastrophic forgetting. 
One important direction is fine-tuning the task-related parameters in soft prompts. 
\citet{yang2022on} proposes to dynamically adjust additional prefixes for each test batch during the testing phase and combine them with the original prefixes to improve LLM robustness, especially against adversarial attacks. 
\citet{Wu_2023_ICCV} leverages CLIP's noisy zero-shot predictions to adjust the task-related prompts in an unsupervised manner, significantly improving CLIP's zero-shot performance. 
Meanwhile, generating high-quality data is also important for improving \robust.
\citet{chen2024understanding} reveals the impact of label noise in LLM applications. They impose regularization in the feature space to mitigate the negative impact of label noise, improving the generalization ability of LLMs. 
\citet{wang2022exploring} explores generative capabilities of LLMs and uses LLMs to generate training corpus, and filters it through the perspective API to ensure low toxicity of the tuning data, thus ensuring \robust. 

Meanwhile, user privacy and fairness problems also receive attention in designing robust PEFT methods. 
\citet{qi2024finetuning} argues that customized fine-tuning will cause the security risks of LLMs, even LLMs have been aligned to reject the generation of harmful content, especially for privacy. 
To ensure that LLM robustness for privacy after fine-tuning, incorporating differential privacy (DP) algorithm into the fine-tuning process is preferred. 
\citet{li2022large} proposes to fine-tune LLMs and employ DP optimization for constructing powerful privacy-preserving LLMs, achieving robust performance to reject generating privacy.
\citet{yu2022differentially} employs new trainable parameters and updates them with differential privacy stochastic gradient descent (DPSGD) while freezing parameters of LLMs, which can effectively fine-tune LLMs while preserving privacy. 
At the same time, directly tuning the attention heads detrimental to privacy work~\cite{zayed2024fairness} is also conducted.

\subsection{Hallucination}
\label{s:knowledge-shift}
As mentioned in Section~\ref{s:ood-detection}, LLMs might be hallucinating when dealing with OOD scenarios. 
Moreover, due to the randomness of sampling strategy, LLMs will generate nonexistence or factual inconsistency content, affecting their reliability and robustness. 
Taking {Fig}.~\ref{f:car-agent} as an example, when a car agent meets extreme weather that is not included in its training data, it may occur hallucination and misidentify blurred lights and shadows in the fog as pedestrians or vehicles. 
The car agent may brake sharply and cause an accident. 
Therefore, hallucination is also one important aspect in OOD robustness of LLMs. 

In general, factual inconsistency is the main concern in hallucination of \robust, which is caused by following reasons: 
1) The generation capability of LLMs, including over-relying on data correlation, over-confidence on generation uncertainty, biased on optimization, etc;
2) Knowledge shift in the data, causing by factual updating, negative impact of data augmentations, and so on. 
To tackle these problems, improving decoding strategies, using better prompts, or employing auxiliary classification models are suggested to alleviate the negative impact of hallucination~\cite{slobodkin2023curious}.
For better decoding strategies, the mainstream is to focus on factual alignment. 
\citet{yuksekgonul2024attention} proposes modeling factual queries as constraint satisfaction problems and uses this framework to study how LLMs interact with factual constraints internally. 
\citet{chuang2024dola} proposes DoLa to align the logits of next-token distribution from the later layers versus earlier layers, thus better surfacing factual knowledge and reducing incorrect fact generation. 
\citet{Ma_2024_CVPR} proposes Vista-LLaMA to address the object hallucination issue by maintaining consistency in the distance between visual and language tokens, especially during the generation of longer texts. 
Meanwhile, using existing knowledge bases to correct the outputs of LLMs is also a promising direction. 
For example, 
\citet{jones2024teaching} proposes to use a synthetic task that hallucinations are easy to elicit and measure for the prefix-tuning. Then, the learned prefix message will be passed to realistic, hard-to-optimize tasks, reducing the hallucinations in LLMs. 
\citet{guan2024mitigating} introduces Knowledge Graph Remodeling (KGR), which combines LLMs with KGs to mitigate factual hallucinations during the inference process by remodeling the initial response draft of the LLM.

Moreover, some works propose incorporating causal reasoning to help LLMs avoid hallucination. 
For example, \citet{wang2021identifying} points out that the main issue with the robustness of LLMs is their reliance on ``spurious correlations'' or ``shortcuts'' between training data and task labels. Thus, they focus on automatically identifying spurious correlations. They first use interpretability methods to extract key markers, then distinguish 'genuine' from 'spurious' markers by analyzing model predictions across multiple corpora, which is useful for mitigating the impact of shortcuts on LLM robustness. 
\citet{yu2024cause} notes that while invoking external knowledge resources to generate richer responses can effectively reduce hallucinations, the inevitable noise in the knowledge can also lead to hallucination problems. Therefore, they explore reasons and future directions for building noise-tolerant methods in knowledge-driven dialogue generation tasks. By analyzing causal relationships and counterfactual reasoning methods, they propose a potential solution by leveraging the interaction between dialogue and knowledge to mitigate hallucination issues. 
\citet{zhou2023causal} address the issue of hallucination in LLMs by proposing Causal-Debias, a framework that mitigates demographic biases and stereotypical associations during fine-tuning. By identifying and intervening on non-causal factors, their method reduces hallucinated biases while maintaining downstream task performance. 
Apart from these work, there also exist other related work~\cite{tian2024finetuning,shang2024ontofact,du2024enhancing,gunjal2024detecting,li2023evaluating,gunjal2024detecting,Jiang_2024_CVPR}.

\textbf{\underline{Summarization.} }
Apart from the general domain transfer and generalization works, the characteristics of LLMs require OOD robustness to focus more on the OOD detection in inputs, PEFT methods in tuning, and Hallucination in outputs. 
Therefore, contrasting diverse positive and negative examples, as well as employing additional LLMs for OOD scenarios detection are the mainstreams. 
For tuning methods, stimulating the few-shot and zero-shot capability of LLMs without tuning too many parameters is the main goal, where constructing high-quality tuning data, designing effective prompts, applying test-time adaption, developing efficient RL-based alignment methods and PEFT methods are current research directions. 
For Hallucination, since factual inconsistency is the main concern, using additional knowledge bases and adding extra fact checking in decoding are the main strategies to reduce hallucination for \robust~improvement.

At the same time, causal inference is also considered in OOD robustness of LLMs. 
Specifically, causality are mainly applied in the LLM alignments and alleviating hallucination, where causal frameworks are employed to identify confounding factors and non-causal factors for reducing biases and improving robustness in OOD scenarios. 
Future research could explore other novel intervention strategies or investigate the causal relationships between data to further enhance the OOD robustness of LLMs.

\section{Robustness Evaluation of LLMs}
\label{s:evaluation}
In this section, we intend to summarize the evaluation for \robust, including new datasets, evaluation methods and benchmarks. 
We have to note that we focus on \robust. Other relevant datasets or benchmarks for LLMs are not covered.


\begin{table}
	\centering
    \small
	\caption{Summarization of existing datasets for evaluating the robustness of LLMs.}
	\label{t:data-statics}
	\begin{tabular}{c|c|c|c}
		\toprule
		Dataset    & Data description  & Data Type & Data statics \\ 
		\midrule
        PromptRobust~\cite{zhu2023promptbench} & \makecell{Measuring LLMs’ resilience to adversarial prompts} & \makecell{character, word, \\sentence, semantic} & 4,788 \\
        \midrule
        CATS~\cite{zhou2020evaluating} & \makecell{Evaluating commonsense ability of LLMs from \\token-level and sentence-level tasks} & Sentences & 323 \\
        \midrule
        UMWP~\cite{sun2024benchmarking} & \makecell{Evaluating LLM hallucination in Question Answering (QA) \\based on the unanswerable math word problem} & math problem & 5,200 \\
        \midrule
        Open-ended QA~\cite{kokaia2023writing} & \makecell{Evaluating domain generalization ability of LLMs \\from closed-book questions to open-book questions} & QA pairs & 1,475 \\
        \midrule
        MultiATIS++~\cite{stickland2022robustification} & \makecell{Focusing on the zero-shot cross-lingual generalization \\and robustness capability on noise inputs} & Multilingualism & 46,421,000 \\
        \midrule
        ROBUSTAPI~\cite{zhong2024can} & \makecell{Evaluating the reliability and robustness of \\code generated by LLMs} & Code & 1,208 \\
        \midrule
        RobuT~\cite{zhao2023robut} & \makecell{Focusing on evaluation on Human-\\Annotated Adversarial Perturbations} & Table QA & 143,477 \\
        \bottomrule
	\end{tabular}
    \vspace{-4mm}
\end{table}

\subsection{Datasets}
\label{s:evaluate-dataset}
To directly evaluate different aspects of \robust~(e.g., open questions, hallucinations, and multilingualism), different datasets are proposed. 
For example, 
\citet{zhou2020evaluating} proposes a dataset CAT to evaluate \robust~in commonsense reasoning tasks. 
\citet{sun2024benchmarking} innovatively develops a dataset Unanswerable Math Word Problem (UMWP) to evaluate hallucinations in LLM question answering based on unanswerable mathematical word problems.
\citet{kokaia2023writing} designs a dataset containing $1,475$ open-ended general knowledge questions, deliberately featuring spelling and grammatical errors, for evaluating domain generalization ability of LLMs (i.e., from closed-book question answering to open-book question answering). 
\citet{stickland2022robustification} curates a robustness dataset for multilingual tasks, including classification/labeling and NLI tasks. 
They generate noisy versions of existing publicly available datasets~\cite{xu2020end, pan2017cross, conneau2018xnli} for evaluating adversarial robustness of LLMs

Besides, other researchers are concerned more about \robust~in downstream applications. 
For example, \citet{zhong2024can} and \citet{shirafuji2023exploring} focus on code generation and develop two programming problem datasets for robustness evaluation. 
\citet{zhao2023robut} creates a new dataset called RobuT, comprising 143,477 examples from WTQ~\cite{pasupat2015compositional}), WikiSQL~\cite{zhong2017seq2sql}, and SQA~\cite{iyyer2017search}, for evaluating \robust~ in able-based question answering. 
Other applications, such as Translation~\cite{jiao2023chatgpt}, toxic speech detection~\cite{hartvigsen2022toxigen}, and RAG evaluation~\cite{chen2309benchmarking} are also proposed. 
Detailed statistics about these datasets can be found in {Table}~\ref{t:data-statics}.

\subsection{Evaluation Methods and Benchmarks}
\label{s:evaluation-method}
LLMs have employed massive parameters to memorize information from vast training corpora. 
Thus, using existing data to evaluate \robust~is risky since we cannot ensure whether the data is used when training LLMs. 
Thus, new benchmarks and metrics are designed for accurately evaluating \robust. 
As summarized in Table~\ref{t:evaluation-methods}, we will introduce these works according to the topology of this paper. 

\subsubsection{Evaluating Adversarial Robustness of LLMs}
As mentioned in Section~\ref{s:adversarial}, Adversarial Robustness mainly focuses on the adversarial noise in the input prompt and output generation.
Thus, synthesizing test data containing various adversarial noises is the dominant framework. 
\citet{zhu2023promptbench} proposes PromptBench, aiming at evaluating the robustness of LLMs against adversarial prompts, which employs various levels of adversarial text attacks across multiple tasks and datasets, generating and assessing numerous adversarial prompts for evaluating LLMs. 
\citet{liu2023trustworthy} uses prompts from the Justice dataset to evaluate the robustness of LLMs in the presence of spelling errors. 
Other similar works, such as SynTextBench~\cite{ko2023robustness}, knowledge-equivalent variants of questions~\cite{zhou2024revisiting}, are also proposed. 
Meanwhile, processing multilingualism and long context is also important for adversarial robustness of LLMs. 
Hence, \citet{dong2023bamboo} introduces a multitask long-context benchmark, BAMBOO, to evaluate LLMs in long-text modeling. This benchmark consists of 10 datasets from five different long-text understanding tasks, covering question answering, hallucination detection, text ranking, language modeling, and code completion across multiple domains.
\citet{ryan2023revisiting} introduces MultiSim benchmark, a collection of 27 resources in 12 different languages, containing over 1.7 million complex-simple sentence pairs for evaluating multilingualism capability of LLMs.

\begin{table}
	\centering
    \small
	\caption{Summarization of evaluation metrics and benchmarks for evaluating the robustness of LLMs.}
	\label{t:evaluation-methods}
	\begin{tabular}{c|c|c}
		\toprule
		Evaluated Capability of LLMs & summarization  & Metrics or Benchmarks \\ 
		\midrule
        \multirow{3}{*}{\makecell{Adversarial Robustness \\evaluation}} 
        & noise or adversarial prompts & \makecell{PromptBench~\cite{zhu2023promptbench}, SynTextBench~\cite{ko2023robustness}, \\knowledge-equivalent questions~\cite{zhou2024revisiting}} \\\cline{2-3}
        & Long-text modeling & BAMBOO~\cite{dong2023bamboo} \\ \cline{2-3}
        & Multilingualism modeling & MultiSim~\cite{ryan2023revisiting} \\
        \midrule
        \multirow{2}{*}{\makecell{OOD Robustness \\evaluation}} 
        & Open domain generalization & ROBUST~\cite{esiobu2023robbie}, SLOG~\cite{li2023slog}\\ \cline{2-3}
        & Reasoning and Inference & \makecell{ROBUSTLR~\cite{sanyal2022robustlr}, JEEBench~\cite{arora2023have}, SAFETEXT~\cite{levy2022safetext}, \\causal-math~\cite{stolfo2022causal}, CLadder~\cite{jin2023cladder}} \\
        \midrule
        \multirow{4}{*}{\makecell{Domain-Related Robustness \\evaluation}} 
        & Code generation & ReCode~\cite{wang2022recode} \\ \cline{2-3}
        & Dialog generation & DGSlow~\cite{li2023white} \\ \cline{2-3}
        & Medical & MedEval~\cite{he2023medeval} \\ \cline{2-3}
        & Science question & UMWP~\cite{sun2024benchmarking}, SCITAB~\cite{lu2023scitab}, Robut~\cite{zhao2023robut} \\
        \midrule
        \makecell{Robustness \\on Biased domains}
        & \makecell{Debiased or fairness \\performance evaluation}& \makecell{NLPositionality~\cite{santy2023nlpositionality}, WinoQueer~\cite{felkner2023winoqueer}, SeeGULL~\cite{jha2023seegull}, \\RTP~\cite{pozzobon2023challenges}, CHBias~\cite{zhao2023chbias}, SocialStigmaQA~\cite{nagireddy2024socialstigmaqa}}\\
        \bottomrule
	\end{tabular}
    \vspace{-4mm}
\end{table}

\subsubsection{Evaluating OOD Robustness of LLMs}
Since LLMs have accessed vast data, it is difficult to construct OOD evaluation data compared with training corpora. 
Therefore, it is natural to collect novel factual data from the real world. 
For example, \citet{esiobu2023robbie} proposes ROBUST, the first benchmark for evaluating open information extraction models in real-world scenarios. Additionally, he refined robustness metrics, using the consistency of model performance across various factions to assess robustness. 
\citet{li2023slog} collects new, complex language expressions to construct a semantic parsing dataset called SLOG for accessing the generalization ability of LLMs. 
Meanwhile, logical reasoning is one important capability that we intend LLMs to master. Moreover, logical or math reasoning data are easy to collect and synthesize. 
Therefore, there are plenty of works focusing on evaluating \robust~on reasoning. 
\citet{sanyal2022robustlr} introduces ROBUSTLR, a diagnostic benchmark based on deductive reasoning, which evaluates the robustness of LLMs to minimal logical edits and different logical equivalence conditions in inputs.
\citet{arora2023have} introduces a challenging problem-solving benchmark for evaluating LLMs called JEEBench. The benchmark was used to analyze the performance of various LLMs in complex logic and mathematical reasoning. 
\citet{levy2022safetext} introduces SAFETEXT, to explore commonsense physical safety issues in NLP. They observe that current LLMs struggle with generating safe text and rejecting unsafe suggestions, highlighting the need for further research. 
\citet{stolfo2022causal} introduces a benchmark to assess the robustness of LLMs, utilizing ASDiv-A~\cite{miao2021diverse}, MAWPS~\cite{koncel2016mawps}, and SVAMP~\cite{patel2021nlp} datasets. This evaluation focuses on causal reasoning factors, including text framing, numerical operands, and operation types.  
\citet{jin2023cladder} also approaches the evaluation of LLMs from a causal perspective by introducing CLadder, a benchmark designed for formal causal inference across all three levels of Pearl’s Ladder of Causation.
This benchmark provides a rigorous testbed to assess and analyze the causal reasoning capabilities of LLMs.
Besides, other works evaluating \robust~in reasoning are also proposed, such as commonsense reasoning generalization~\cite{zhou2020evaluating,shen2024generalization}.

\subsubsection{Evaluating Domain-Related Robustness of LLMs}
Apart from the general capabilities of \robust, researchers are also concerned about the robustness performance of LLMs in specific downstream areas. 
For example, in code generation, \citet{wang2022recode} introduces ReCode, a benchmark for evaluating the robustness of LLMs in code generation, which generates perturbations in code docstrings, functions, syntax, and format, including character-level and word-level insertions or transformations for comprehensive evaluation. 
In dialog generation, \citet{li2023white} proposes a novel DGSlow to use white-box attacks to verify the robustness of LLMs in dialog generation.  
\citet{he2023medeval} introduces MedEval, a multi-level, multi-task, and multi-domain medical benchmark dataset designed to advance the development of LLMs in the medical field.
Besides, other areas, such as scientific questions~\cite{lu2023scitab,sun2024benchmarking}, Table QA~\cite{zhao2023robut}, are also developed. 

Meanwhile, LLMs are expected to be robust when facing bias and toxicity situations, which is essential in application.
In this area, \citet{santy2023nlpositionality} introduces a framework called NLPositionality, which quantifies design biases in NLP datasets and models by collecting annotations from diverse populations. 
\citet{felkner2023winoqueer} introduces a benchmark WinoQueer, to detect biases in LLMs concerning the LGBTQ+ community. 
Meanwhile, there also exist benchmarks related to different bias-type, such as gender~\cite{hada2023fifty}, racial~\cite{deas2023evaluation}, stereotype~\cite{jha2023seegull}, toxicity~\cite{qi2023preserving,pozzobon2023challenges}, and social bias~\cite{zhao2023chbias,nagireddy2024socialstigmaqa}.

\textbf{\underline{Summarization.} }
To evaluate the \robust~comprehensively, designing novel datasets is the optimal direction. 
Thus, plenty of challenging datasets are designed to evaluate \robust~on general capabilities (e.g., question answering, knowledge extraction, generation, etc). 
Moreover, constructing downstream application-related datasets is also an important direction.

Meanwhile, other benchmarks and evaluation methods are also developed to evaluate the specific aspect (e.g., the capability of processing long context) of \robust. 
As for causal reasoning capability, researchers concentrate more on how well LLMs understand the relationship between causality and outcomes.
These benchmarks reflect a growing awareness of the need to incorporate causal analysis into the evaluation of \robust, ensuring that LLMs can solve complex tasks beyond mere pattern recognition.
This direction still has many aspects worth exploring, and we believe that causality can facilitate the evolution of LLMs.

\begin{figure}
    \centering
    \includegraphics[width=0.9\linewidth]{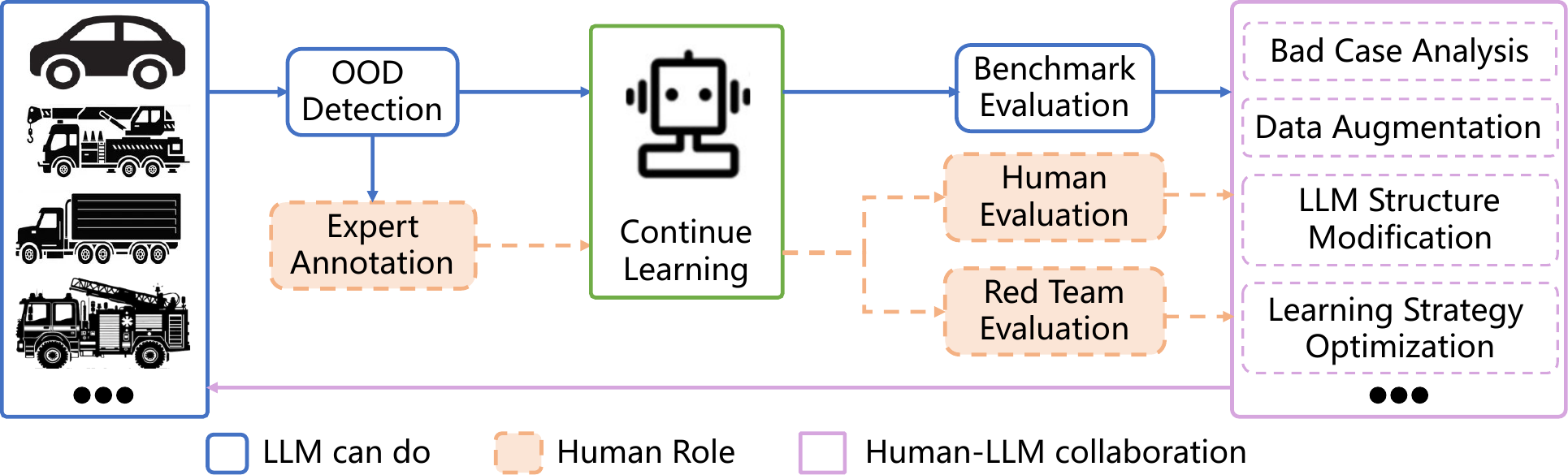}
    \caption{An example of the human-in-the-loop framework for continually improving the robustness of LLMs.}
    \label{f:continue-learning}
    \vspace{-4mm}
\end{figure}

\section{Discussion and Future Directions}
\label{s:discuss-future}
In this section, we mainly discuss the human impact on the research of \robust. After that, we present some possible promising direction from the adversarial robustness, OOD robustness, and evaluation perspectives.

\subsection{Discussion about Human Role in \robust}
\label{s:discussion}
Different from traditional neural networks, the impact of human is essential in training LLMs, such as using human feedback as the reward of reinforcement learning in Alignment~\cite{kaufmann2023survey}. 
When discussing \robust, we argue that it is important to consider the human function and impacts. 
Considering the capabilities of LLMs, we discuss the human impacts for the following aspects: 
\begin{itemize}
    \item[1. ] \textit{Annotator. } Though LLMs are pre-trained with existing corpora, it is still important to provide high-quality labeled data, which humans are the best option to annotate these corpora. Moreover, even if synthesizing data has become a hot topic recently, there already exist works to prove the negative impacts on \robust. Thus, humans still play an integral role in providing high-quality data. 

    \item[2. ] \textit{Experts. } When applying LLMs to downstream areas, it is necessary to incorporate domain knowledge for better adapting. Therefore, humans can act as experts to provide formulated domain requirements, structured domain knowledge, pre-defined domain risks, and so on. 
    Then, LLMs can learn from experts to improve their robustness. 
    In this scenarios, Humans act as the teacher to guide LLMs, which can ease the process of domain adaptation. 
    
    \item[3. ] \textit{Red Team. }
    In adversarial robustness of LLMs, LLMs are expected to be robust when dealing with adversarial prompts. 
    Therefore, humans can play as red team to design adversarial prompts and attacking methods to explore the boundaries of LLMs, which can evaluate the LLM capability and provide detailed suggestions for further improvement.

    \item[4. ] \textit{Evaluator. }
    Since LLMs have demonstrated impressive performance over various scenarios, it is challenging to evaluate \robust~comprehensively and accurately. 
    In this situation, humans can act as the evaluator to evaluate and verify the generation of LLMs. 
    Moreover, human evaluators can provide detailed comments of the generation, including strengthens and weaknesses. 
    Furthermore, in some situations that evaluation metrics are hard to design, human evaluators are also the best option. 

\end{itemize}
Moreover, we have provided an example of the human-in-the-loop frameworks for continually improving the robustness of LLMs, which have been illustrated in Fig.~\ref{f:continue-learning}. 
In the continue learning scenario of car agent in this figure, we first ask LLMs to identify uncertainty examples that may be out of their capability. 
Then, we collect these examples and ask experts to evaluate them and annotate whether these examples belong to the distribution of tuning data. 
Next, annotated data will be added to the tuning data or used annotated data to further update the tuning strategies of LLMs. 
The following is the updating process of LLMs. 
By using this framework, humans can provide assistance for the improvement of \robust. 
Other frameworks share a similar motivation.

\subsection{Future Direction}
\label{future}
Existing works have improved the \robust~from different perspectives. 
However, there are still many ongoing problems of \robust~remaining unsolved. 
In response, we present some possible promising directions that deserve more efforts from short-term, medium-term, and long-term perspectives. 

\underline{\textit{Short-Term Goal. }}
For the next two years, the main target is improving the capability and efficiency of LLMs when dealing with various adversarial prompts and OOD scenarios. 
Thus, long context modeling and multi-modal prompts are two main directions in improving adversarial robustness of LLMs. 
Novel sparse attention mechanisms, memory units, and alignment strategies are promising for LLM capability improvement. 
Meanwhile, wider applications of LLMs require PEFT defense strategies for LLMs to defend against various attacks without high-cost tuning. 
For OOD robustness, we believe the most urgent task is to develop standardized OOD evaluations, so that various LLMs can be evaluated and compared under a fair and comprehensive framework. 
After that, incorporating human efforts to design effective data augmentations is important to satisfy the massive data requirements of LLMs. 
Meanwhile, domain-related PEFT strategies are also important for improving the capability of LLMs in specific areas. 
For robustness evaluation, there is still a lack of general and uniform benchmarks for LLM evaluation. 
Moreover, impressive LLM capabilities and various downstream applications make it hard to evaluate LLMs comprehensively. 
Therefore, it is also promising to take humans as the judges to finish the evaluation process, which is also known as human-in-the-loop evaluations.

\begin{table}
	\centering
    \small
	\caption{Summarization of future directions of \robust~at different stages.}
	\label{t:future}
	\begin{tabular}{c|c|c|c}
		\toprule
		Stage & \makecell{Adversarial \\Robustness}  & \makecell{OOD \\Robustness} & \makecell{Robustness \\Evaluation} \\ 
		\midrule
        \makecell{Short-Term \\Goal} 
        & \makecell{Long context processing, \\Multi-modal prompts, \\PEFT Defense Strategies} 
        & \makecell{Standard OOD evaluation, \\Data Augmentation, \\Domain-related PEFT} 
        & \makecell{Uniform Benchmarks, \\Human-in-the-loop \\Evaluation} \\
        \midrule
        \makecell{Medium-Term \\Goal} 
        & \makecell{Self-Supervised Adversarial Training, \\Multi-modal Attack Transfer, \\Explainable and Transferrable Defense} 
        & \makecell{Novel OOD Detection strategies, \\ Causal Invariant Learning, \\General LLM design} 
        & \makecell{Excessive Ability Test, \\Explainable Evaluation, \\Vulnerability Localization} \\
        \midrule
        \makecell{Long-Term \\Goal} 
        & \makecell{Theory of Adversarial Robustness, \\Causal-Inspired Learning paradigms, \\Human-LLM Cooperation} 
        & \makecell{Life-long Adaptive Generalization, \\World Models, \\Human-LLM Cooperation} 
        & \makecell{Self-Assessment agents, \\Social Impact Assessment, \\Human-LLM Cooperation} \\
        \bottomrule
	\end{tabular}
    \vspace{-5mm}
\end{table}

\underline{\textit{{Medium-Term Goal. }}}
For the next five years, we believe more efforts will be conducted in advanced learning or tuning paradigms. 
E.g., self-supervised adversarial training methods will be developed to improve the \robust~when dealing with diverse noise prompts. 
Better attack and defense strategies are also gained more and more attentions, such as multi-modal attack transfer methods, transferrable and explainable defense methods. 
All these methods will help to figure out more inner mechanism of LLMs. 
For OOD robustness, LLMs are expected to be aware of their capability boundaries. 
Therefore, novel OOD detection methods will be developed and verified. 
Moreover, to further maintain the performance of LLMs in open scenarios, causal invariant learning paradigms and general LLM design will become the hot research directions to help LLM learn the invariant knowledge and features about the world. 
Moreover, the robustness evaluation will provide detailed and precise analysis about LLMs, such as excessive ability test, explainable evaluation, and vulnerability localization. 
These evaluation frameworks will provide detailed feedback for humans for further human-LLM cooperation.

\underline{\textit{{Long-Term Goal. }}}
Apart from the short-term goal and medium-term goal, the ultimate goal of LLMs is to realize Artificial General Intelligent (AGI). 
Therefore, the long-term goal of \robust~is to realize the robust, stable, and trustworthy performance when dealing with massive real-world scenarios. 
For example, the theory of adversarial robustness and causal-inspired learning paradigms will be developed to explain and boost the LLM capability of tackling noise prompts. 
Moreover, life-long adaptive generation and world models will become the mainstream when dealing with real-world OOD scenarios. 
Furthermore, self-assessment, social impact, and other real-world impacts will be considered more when evaluating the capability of LLMs. 

Besides, we argue that human-LLMs cooperation will play a pivotal and indispensable role in the \robust~development. 
We believe that it is impractical that LLMs will replace humans in the real world. 
As an intelligent tool, LLMs are supposed to provide assistance to humans and help them realize their intelligence more conveniently, efficiently, and comprehensively. 
Meanwhile, with the collaboration of human intelligence and AI, we will be able to build a better world. 
Therefore, human-LLMs cooperation will be a long-term goal of \robust~as well as LLMs development.

\section{Conclusion}
In this paper, we focus on \robust~and provide a systematic review about the cutting-edge researches. 
Following the definition of \mlrobust, we first give a formal definition of \robust~based on the characteristics of LLMs. 
For adversarial robustness, we summarize the influential research work from \textit{noise prompt} and \textit{Noise Decoding} perspective. 
For OOD robustness, we present the representative work from \textit{OOD Detection}, \textit{PEFT Methods}, and \textit{Hallucination} aspects. 
Meanwhile, we also provide evaluation datasets and benchmarks for \robust. 
Apart from summarizing existing works, we also discuss and highlight future opportunities of \robust. 
Moreover, we provide a github repo to organize related works in this survey for easy access. 
Considering the rapid development of LLMs, our paper cannot cover all works related to \robust, since a large number of novel methods are proposed in each year. 
We hope our survey can provide a quick understanding and overall framework perception of \robust~for readers, so that they can quickly capture their interest.



\bibliographystyle{ACM-Reference-Format}
\bibliography{8-reference}










\end{document}